\documentclass[twoside]{article}

\usepackage[preprint]{aistats2027}
\newcommand{\corresponding}{\textsuperscript{*}}
\usepackage{url}
\usepackage{algorithm}
\usepackage{algorithmic}
\usepackage{amsmath}    % 数学公式、\mathbf、tanh、矩阵符号
\usepackage{amssymb}   % \mathbb{R}、半正定符号 \succeq
\usepackage{booktabs}  % 表格三线表 \toprule \midrule \bottomrule
\usepackage{pifont}    
\usepackage{newfloat}
\usepackage{listings}
\usepackage{xcolor}
\usepackage{graphicx}
\usepackage{multirow}

\begin{document}

\runningtitle{VIA: Tailored Attention Matters for Science}
\runningauthor{Wang Rui}



\twocolumn[

\aistatstitle{\textbf{\textcolor{red}{V}}ariational-\textbf{\textcolor{red}{I}}sing-\textbf{\textcolor{red}{A}}ttention: Tailored Attention Matters for Science}

\aistatsauthor{ Wang Rui \corresponding, }

\aistatsaddress{ Tsinghua University\\
    Department of Chemical Engineering\\
    r-wang18@tsinghua.org.cn\\
    ORCiD: 0000-0002-2362-3214\\
    {\small * Corresponding author.} } ]

\begin{abstract}
 Attention enables context modeling \textit{via} query-key scoring with softmax normalization. Driven by industrial long-context demands, mainstream research has converged toward sparsity and efficiency—yet softmax's independence assumption persists. For scientific tasks unburdened by long-token constraints, however, richer structured coupling may often be essential, making tailored attention both viable and more appropriate. To this end, we propose Variational-Ising-Attention (VIA), which augments softmax normalization with an interacting Ising model; attention patterns emerge from learnable pairwise couplings \textit{via} variational mean-field inference, extending attention from a ranking over isolated items to a collective state over interacting entities. We instantiate VIA on retrosynthesis reaction center prediction and, as a controlled internal ablation, on protein residue contact prediction — two structured prediction tasks governed by cooperative constraints: cooperative bond-breaking for retrosynthesis and inter-residue interactions for protein contact prediction. Comprehensive experiments across model variants, coupled with mechanistic analyses, demonstrate that VIA substantially outperforms standard softmax attention. More broadly, our findings suggest that for scientific problems, the optimal solution is not general-purpose efficiency, but appropriately tailored attention aligned with intrinsic domain structure. This work provides a theoretically grounded and empirically validated instantiation of this paradigm.
\end{abstract}

\section{Introduction}

\begin{figure*}[t]
\centering
\includegraphics[width=0.85\textwidth]{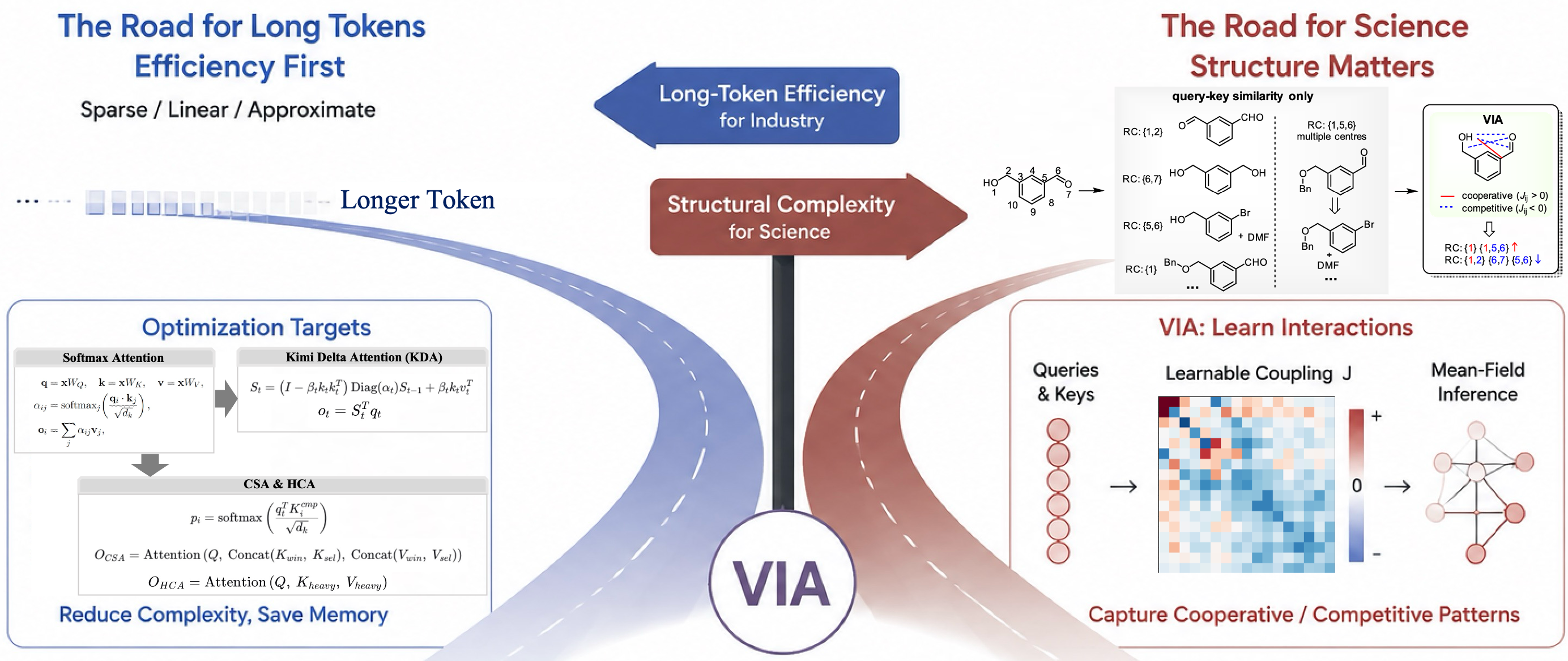}
\caption{A Distinct Direction for Science}
\label{fig1p}
\end{figure*}

Attention mechanisms, introduced by the Transformer architecture (Vaswani et al., 2017), are foundational in modern neural architectures. In industry, the drive for long-context modeling has focused on efficiency \textit{via} sparsity and linear approximations, as exemplified by sparse attention variants such as Longformer (Beltagy et al., 2020), BigBird (Zaheer et al., 2020), and more recent industrial systems like DeepSeek-V4 (DeepSeek-AI, 2025; Wang et al., 2026) and Kimi Delta Attention (KDA) (Moonshot AI, 2024) (Blue Road in Figure~\ref{fig1p}). Scientific tasks, however, pose a different challenge: sequence lengths are moderate, but structural complexity is high---entities interact, compete, and cooperate. Standard softmax attention scores queries and keys independently and normalizes in one shot, lacking mechanisms for inter-position correlations.

Prior attempts to address this fall into two categories. Energy-based variants (Hopfield networks, Ramsauer et al., 2021), Energy Transformers (Hoover et al., 2023), AttnBM (Ota and Karakida, 2023)) remain non-interacting ($\mathbf{J}=0$), capturing query--key but not position--position interactions. Sigmoid attention (Ramapuram et al., 2025) shares this limitation. Approaches that introduce couplings---QAMA (Du et al., 2025) derives them as fixed functions of scores, underperforming softmax; Spin-Model Transformer (Bal, 2023) learns with mean-field but reports no gains.

Within scientific domains, parallel efforts have explored task-specific attention biases to inject structural inductive biases. As a pioneering work in geometry-aware molecular attention design, Uni-Mol (Zhou et al., 2023) injects 3D geometric priors through a frozen Gaussian kernel and achieves notable empirical gains (Wang et al., 2024; Luo et al., 2024). Nevertheless, it relies on computationally expensive pre-computed atomic coordinates, adopts a data-agnostic fixed kernel, and incurs substantial persistent memory overhead.

We propose \textsc{Variational-Ising-Attention} (VIA), which augments attention with a learnable pairwise coupling matrix capturing cooperative/competitive interactions. Attention weights are obtained with iterative mean-field inference while retaining the softmax operation. VIA learns couplings directly from data and remains lightweight through diverse parameterization.

We validate VIA on retrosynthesis reaction center prediction, with protein residue contact prediction as a supplementary internal ablation; mechanistic analyses reveal that VIA flattens the energy landscape, expands representational capacity, and learns chemically meaningful patterns.

Through systematic comparison of deployment strategies, backbone deployment substantially outperforms terminal-only attachment, while both consistently outperform standard softmax. These findings suggest that for tasks with intrinsic structural coupling, modeling capacity stems from domain-tailored interaction mechanisms rather than increasingly efficient general-purpose attention—a principle VIA instantiates through learnable structural couplings (Red Road in Figure~\ref{fig1p}).

\section{VIA setup}

We introduce \textsc{Variational-Ising-Attention} (VIA), an interaction-mediated attention mechanism that augments standard self-attention with a learnable signed pairwise coupling matrix. The formulation is inspired by Ising-type energy models (McCoy and Wu, 1973; Sherrington and Kirkpatrick, 1975; Baxter, 1982) and implemented \textit{via} damped mean-field-inspired updates. We study two deployment regimes: Terminal VIA (TV) and Backbone/Full VIA (FV).

We start from an Ising-type energy:
\begin{equation}
    E(\mathbf{s};\mathbf{h},\mathbf{J})
    =
    -\sum_j h_j s_j
    -\frac{1}{2}\sum_{j\neq k}J_{jk}s_js_k,
\end{equation}

where $s_j\in\{-1,+1\}$ denotes a binary interaction state,
$h_j$ is the unary local field, and $\mathbf{J}$ is a signed pairwise
coupling matrix. Positive and negative couplings represent cooperative
and competitive interactions, respectively.

Given query and key representations, we compute the scaled
query-key compatibility matrix
\begin{equation}
    \mathbf{S}
    =
    \frac{\mathbf{Q}\mathbf{K}^{\top}}{\sqrt{d_k}},
\end{equation}
and use its diagonal as the unary local field:
\begin{equation}
    h_j=S_{jj}.
\end{equation}

Thus, the local compatibility initializes the interaction state, while $\mathbf{J}$ propagates relational information across positions.

Exact inference over the Ising distribution is intractable (Jain et al. 2018), and Monte Carlo sampling yields high-variance gradients that destabilize end-to-end training (Mohamed et al. 2020). We therefore adopt deterministic mean-field updates, which are differentiable and low-variance (Peterson \& Anderson 1987). For TV, which models reaction-center decisions as independent binary variables, we use a coordinate-wise sigmoid update
\begin{equation}
    m_j^{(t)}
    =
    \sigma\left(
    \frac{
    h_j+\gamma\sum_{k\neq j}J_{jk}m_k^{(t-1)}
    }{T}
    \right).
\end{equation}

This formulation is consistent with the coordinate‑wise sigmoid/tanh
form arising from factorized mean‑field inference for binary Ising
variables (Parisi, 1988; Bishop, 2006;
Wainwright and Jordan, 2008; Jain et al., 2018).

For FV, the interaction state serves as an internal normalized routing representation. We therefore use a simplex-constrained update:
\begin{align}
    \tilde{m}_j^{(t)}
    &=
    \operatorname{softmax}_j
    \left(
    \frac{
    h_j+\gamma\sum_{k\neq j}J_{jk}m_k^{(t-1)}
    }{T}
    \right),\\
    m_j^{(t)}
    &=
    \lambda m_j^{(t-1)}
    +(1-\lambda)\tilde{m}_j^{(t)}.
\end{align}

Unlike the binary-Ising mean-field state, this normalized interaction
state satisfies $\sum_jm_j=1$ and is therefore not interpreted as the
binary-Ising marginal $p(s_j=+1)$. Instead, it provides a normalized
interaction context for attention routing.

After interaction-state inference, the coupling matrix is used to
construct the attention field:
\begin{equation}
    e_{ij}
    =
    h_i+\gamma J_{ij}m_j^{(K)}.
\end{equation}

Because $h_i$ is constant with respect to the normalization index $j$,
it cancels from the final row-wise softmax. The effective coupling-aware
attention is therefore
\begin{equation}
    \alpha_{ij}
    =
    \operatorname{softmax}_{j}
    \left(
    \frac{\gamma J_{ij}m_j^{(K)}}{T}
    \right),
\end{equation}
followed by value aggregation
\begin{equation}
    Z_i=\sum_j\alpha_{ij}V_j.
\end{equation}

Hence, VIA performs interaction-mediated routing via $h\rightarrow m^{(K)}\rightarrow Jm^{(K)}\rightarrow \alpha$, rather than applying softmax directly to the full pairwise compatibility matrix (tensor shapes for both TV and FV in Appendix~A1.1). We investigate four parameterizations of $\mathbf{J}$; details follow. 

\section{Retrosynthesis Reaction Center Prediction}

We mainly evaluate VIA on retrosynthesis reaction center prediction, a task that inherently requires coordinated decision‑making across multiple bonds (Wan et al., 2022; Zhong et al., 2023; Chen and Jung, 2021; Yan et al., 2020; Wang et al., 2026). Below we describe the task, data, model configurations, and evaluation protocols; implementation details (hyperparameters, architectural variants, and diagnostic formula derivations) are deferred to the Appendix A2.

\subsection{Experimental Setup}

Retrosynthesis prediction aims to identify precursor molecules for a given target product. A critical sub-task is \textbf{reaction center identification}: predicting the set of atoms that participate in bond-breaking events, formulated as binary classification over atom pairs (see Figure 1a, red route). We divide the task into two regimes based on reactive site complexity: a \textbf{single-center regime} (1-RC and weakly coupled 2-RC cases, where bond breaking is confined to a localized motif) and a \textbf{multi-center regime} (remaining 2-RC and all $\ge$3-RC cases, where breaking proceeds at multiple independent sites requiring coordinated modeling). Multi-center reactions are particularly challenging: missing any single breaking bond invalidates the entire retrosynthesis path, and decisions across sites are coupled \textit{via} electronic and steric interactions—precisely the structure VIA provides.

\subsubsection{Data and Features}

We use USPTO-50k (Lowe et al., 2012), a standard benchmark of single-step organic reactions. After RDKit parsing, 47{,}671 of the nominal 50{,}000 reactions are retained. Inputs are product SMILES parsed into 30-dimensional atom features encoding local chemical environments (atomic number, degree, hybridization, aromaticity, ring membership, formal charge). These features are purely local and contain no pre-computed global structure, so the model must learn inter-atom dependencies entirely from data (Figure 2a). To prevent information leakage, reaction labels are derived from atom mappings between products and reactants, but neither the mapping itself nor reaction-type annotations are exposed to the model. We follow the standard 8:1:1 train/validation/test split. Molecules are padded to $N_{\max}=100$, with a binary mask excluding padding from all computations; the coupling matrix $\mathbf{J}$ has shape $N_{\max}\times N_{\max}$, and all diagnostics are computed only on the $N_{\text{actual}}\times N_{\text{actual}}$ submatrix of real atoms.

\subsubsection{Architectures and parameterizations}

All models share the same high-level structure: an encoder processing atom features, followed by a prediction head outputting reaction center probabilities per atom pair. We consider three configurations: a \textbf{Softmax Baseline} with standard multi-head softmax attention (2 layers, 8 heads, hidden size 512); \textbf{TV (Terminal Deployment)}, which uses a softmax encoder followed by a single static Ising attention head at the terminal prediction layer, so the physical prior does not participate in feature extraction; and \textbf{FV (Backbone Deployment)}, which uses VIA instead of softmax attention in all encoder layers (same depth and width), allowing the Ising dynamics to shape features from the first layer onward (Figure 2b). The key distinction is not depth (all use 2 layers) but \textit{where} the Ising structure is applied. For each VIA-based architecture (TV and FV), we explore four parameterizations of the coupling matrix $\mathbf{J}$ (Figure 2c): Original (full $N^2$), LowRank (low-rank factorization), Hyper (input-conditioned \textit{via} hypernetwork), and PairMLP (per-pair MLP). In \textbf{FV}, LowRank uses a \textit{fixed} learnable matrix $\mathbf{U}\in\mathbb{R}^{N\times r}$ with $r=64$ (independent of input), ensuring positive semi-definiteness and stable convergence; in \textbf{TV}, LowRank instead uses a hypernetwork to dynamically generate $\mathbf{U}$ conditioned on the input features, with rank $r=32$. Complete definitions and parameter counts are provided in Appendix A2.

\subsubsection{Evaluation Metrics}

We formulate reaction center prediction as atom-level binary classification: the model outputs a reactivity probability $P(\text{atom}_i \in \text{RC})$ for each atom, with binary labels $\{0,1\}$ where 1 denotes a reaction center atom. We adopt \textbf{dynamic top-$k$ exact-match} as our primary metric: for each sample with $k$ ground-truth centers, the prediction is correct iff the top-$k$ predicted atoms exactly match the ground-truth set. This directly reflects single-step retrosynthesis success. We additionally track \textbf{atom-level F1} as a complementary training proxy, detailed in Appendix A2.

\begin{figure*}[!t]
\centering
\includegraphics[width=0.88\textwidth]{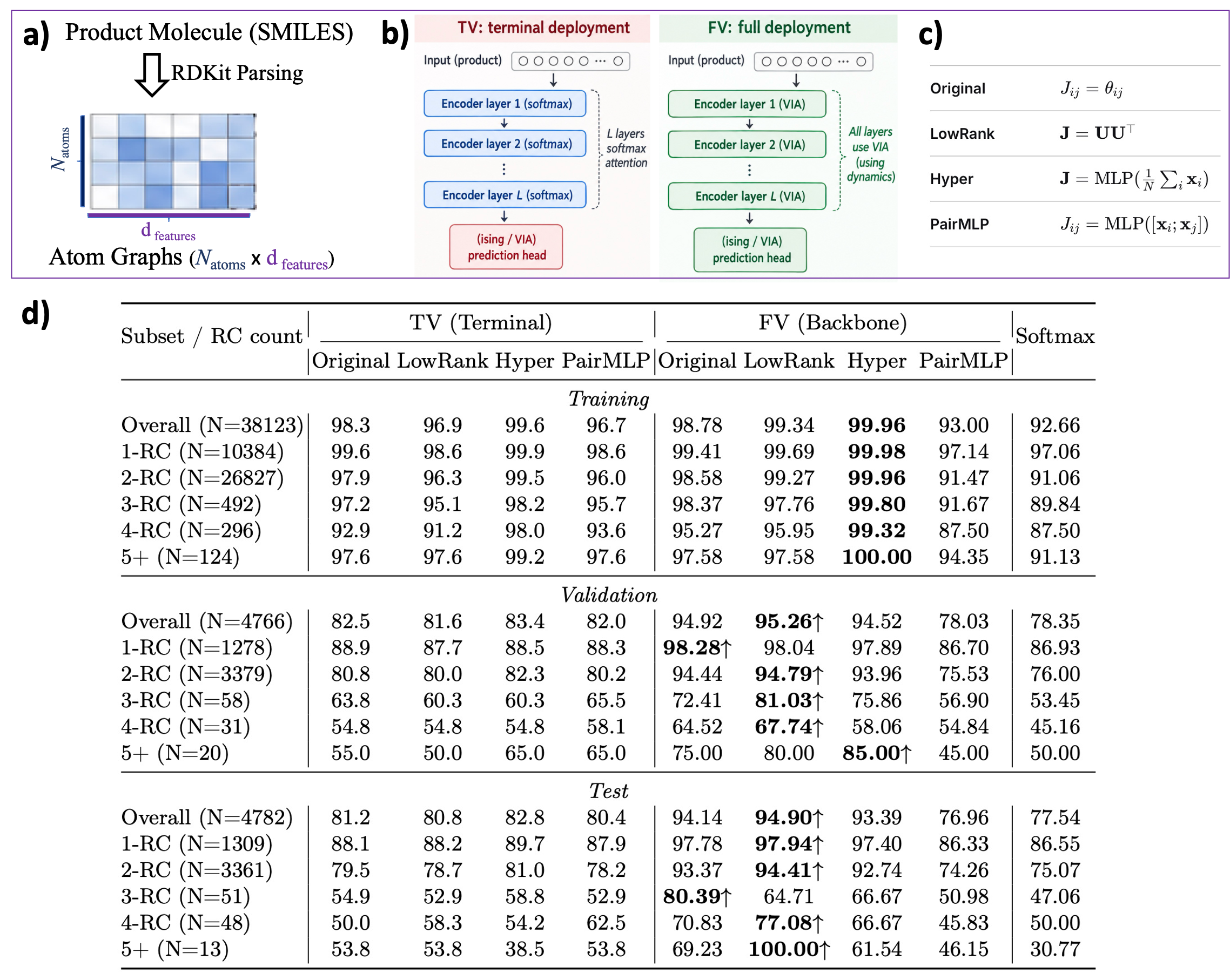}
\caption{Model setup and performance of VIA}
\label{f2pp}
\end{figure*}

\subsubsection{Diagnostic Tools}

To investigate why VIA works and why deployment location matters, we apply three diagnostics to trained models: the geometric frustration index, effective rank, and $\mathbf{J}$ matrix visualization. Their definitions are given in Appendix~A2.4, and their use is described in Section~3.2.2. All diagnostics are computed on the real-atom submatrix, excluding padding.

\subsection{Results}

We empirically validate VIA on retrosynthesis reaction center prediction through ablations over coupling parameterizations, inference schemes, and deployment regimes. All main experiments adopt variational mean-field inference; the Gibbs-sampling alternative is reported in Appendix~A2.11. Below we present training dynamics, held-out benchmark performance, and mechanistic diagnostics. All quantitative results use the dynamic top-$k$ exact-match protocol defined in the Experimental Setup, with per-subset breakdowns in Figure~2d, with training dynamics showed in Appendix A2.3.

\subsubsection{Benchmark Performance}

We evaluate all models on the held-out test set under the unified dynamic top-$k$ exact-match criterion, with full numerical breakdowns by reaction center count across training, validation, and test splits summarized in Figure 2d.

Consistent with the training dynamics observed above, FV backbone variants exhibit strong generalization performance. On the training set, FV-Original, FV-LowRank and FV-Hyper achieve overall exact-match accuracy in the range of 98.78\%–99.96\%, showing only modest gains relative to their TV terminal counterparts. More notably, on the validation set they reach 94.52\%–95.26\% overall accuracy, with substantially reduced overfitting and markedly better generalization compared to TV variants.

On the full test set of 4,782 reactions, backbone deployment of VIA delivers consistent and substantial improvements over both terminal deployment and the softmax baseline. FV-LowRank achieves the highest overall exact-match accuracy at 94.90\%, followed by FV-Original at 94.14\%. In comparison, the best-performing TV variant (TV-Hyper) reaches 82.8\%, whereas the standard softmax baseline yields 77.54\% (FV) and 79.5\% (TV), respectively. This performance advantage is stable across data splits, confirming that structured pairwise couplings embedded in the backbone provide a generalizable inductive bias rather than overfitting to training data.

The advantage of VIA generally becomes larger for multi-center reactions, although the improvement is not strictly monotonic across individual reaction-center counts. For single-center reactions, FV-LowRank achieves 97.94\% accuracy, already outperforming the softmax baseline and indicating that structured couplings can benefit localized reactive motifs. For multi-center reactions, the performance gap becomes substantially larger, reaching 27.0 percentage points for reactions with four or more reaction centers. These results are consistent with the hypothesis that explicitly modeling cooperative and competitive inter-atomic interactions is particularly beneficial when multiple coordinated bond-breaking sites must be identified. The relatively small 3-center stratum ($n=51$) also introduces substantial sampling uncertainty, so we do not interpret the trend as strictly monotonic.

A reaction-class and molecular-size stratification further suggests that VIA's advantage holds across all ten reaction classes, with the largest gaps on deprotection, reduction, and oxidation reactions where cooperative bond-breaking is most pronounced; details are provided in Appendix~A2.8.

To our knowledge, existing RC prediction protocols on USPTO-50K report SOTA comparable to our terminal-deployed TV variants  (Wang et al., 2026), since their information interaction logic is equivalent to a terminal head. Our FV backbone variants substantially surpass this level, showing that the decisive factor is where structural biases are deployed, not whether they are introduced.

\subsubsection{Mechanistic Insights}

\begin{figure*}[!t]
\centering
\includegraphics[width=0.88\textwidth]{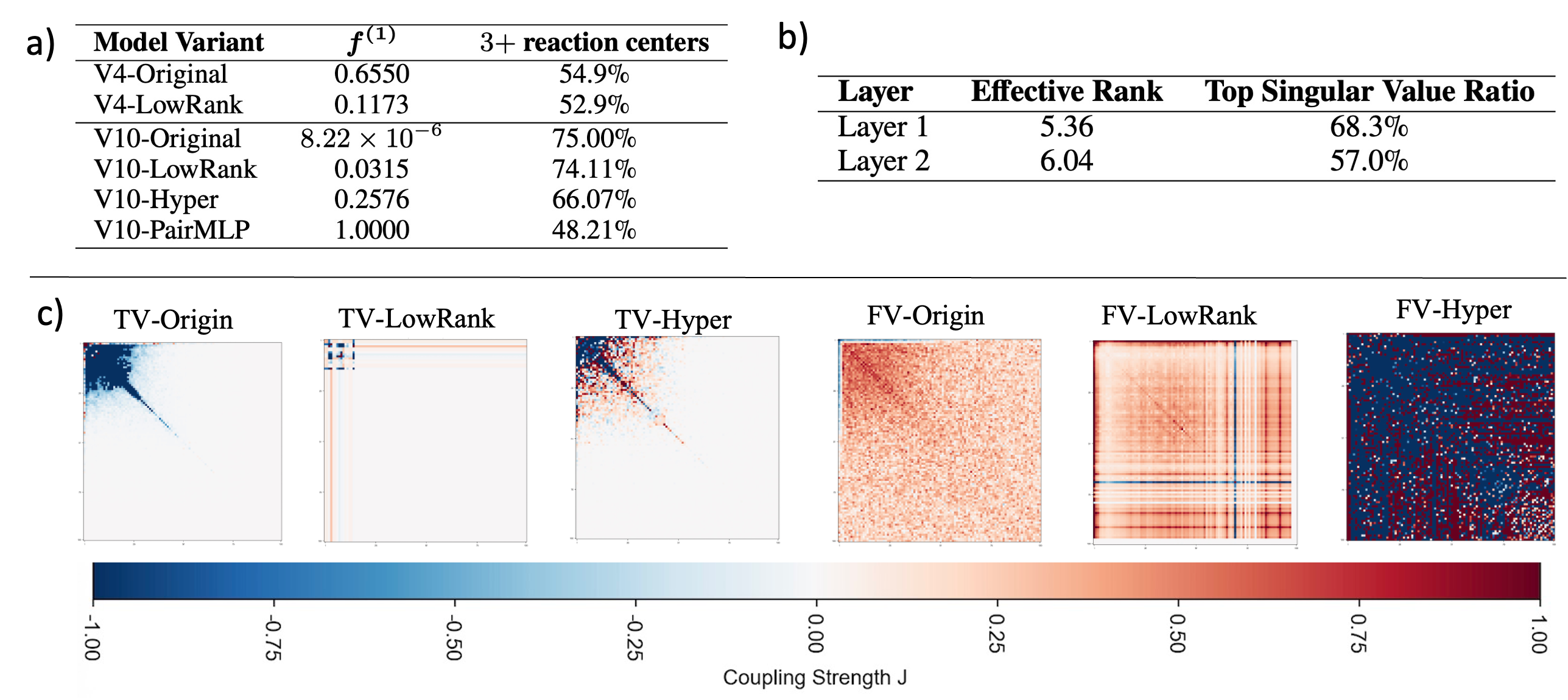}
\caption{Mechanistic Analysis Results}
\label{f3p}
\end{figure*}

To elucidate the origins of VIA’s empirical advantages, we conduct three core diagnostic analyses of the learned coupling matrix $\mathbf{J}$: geometric frustration index, effective rank evolution, and coupling pattern visualization. 

\paragraph{Geometric Frustration}
The geometric frustration index quantifies triadic conflicts in the learned coupling matrix $J$. Lower frustration is associated with a smoother coupling landscape under the corresponding Ising interpretation. Here, we use this quantity as a structural diagnostic of the learned interactions rather than as a direct measure of predictive performance or convergence. Numerical values for first-layer frustration are reported in Figure 3a.

We observe that backbone integration dramatically reduces first-layer frustration. For the Original parameterization, frustration drops from 0.6550 in TV to $8.22\times10^{-6}$ in FV—a reduction of nearly five orders of magnitude. For LowRank, frustration falls from 0.1173 to 0.0315. This correlation suggests that backbone integration enables the model to learn representations with more favorable coupling structures. In the terminal configuration, by contrast, the Ising head operates on features produced by a softmax encoder never exposed to coupling constraints, and the single terminal layer has limited capacity to resolve inherited triadic conflicts.

Across parameterizations, PairMLP exhibits maximum frustration ($f=1.0$) in both deployment schemes, indicating complete saturation of conflicting triadic interactions. This suggests that per-pair independent computation breaks the global coupling structure required for stable Ising dynamics, leading to gradient degradation and poor task performance. The Hyper variant shows intermediate frustration and intermediate performance, consistent with unconstrained hypernetwork generation introducing spurious pairwise correlations.

We further compare terminal-layer frustration between correctly and incorrectly predicted samples. Hyper is the only variant where incorrect predictions are systematically associated with higher frustration ($\Delta=-0.0455$), meaning its noisy coupling patterns disproportionately degrade performance on difficult samples. Original and LowRank show negligible frustration gaps, indicating their smooth energy landscapes handle easy and challenging cases with equal reliability.

For LowRank, a closer examination reveals a counter-intuitive positive correlation between frustration and accuracy under its PSD-constrained factorization, suggesting that the diagnostic proxies for interaction diversity in this setting; full stratification results are provided in Appendix A2.6.

\paragraph{Effective Rank Evolution}
For the best-performing FV-LowRank variant, we track the effective rank of $\mathbf{J}$ across encoder layers to measure how representational capacity evolves during feature extraction. Results are summarized in Figure 3b.

Effective rank expands from 5.36 in layer 1 to 6.04 in layer 2, while the top singular value ratio decreases from 68.3\% to 57.0\%. This controlled rank expansion is functionally significant: An effective rank above 5 indicates that the coupling matrix maintains multiple independent interaction modes, which is consistent with the model's ability to handle multi-center reaction patterns. The model gradually disperses its spectral weight across more independent interaction modes, building capacity for diverse cooperative reaction patterns rather than overfitting to a small set of dominant motifs. This progressive capacity building is only possible when couplings participate in every layer of feature learning; a single terminal Ising layer cannot incrementally expand representational rank in this way.

Cross-model stratification by effective rank (Appendix A2.6) further reveals that moderate rank \([8,12)\) optimizes performance for most parameterizations, while LowRank's monotonic decrease highlights the regularization effect of its built-in bottleneck.

\paragraph{Coupling Matrix Visualization}
Visual inspection of the learned $\mathbf{J}$ matrices reveals a clear correlation between global structural organization and task performance. FV-LowRank exhibits the most interpretable structure: strong diagonal self-reinforcement and distinct block-structured off-diagonal bands, corresponding to chemically meaningful cooperative (positive) and competitive (negative) relationships between classes of atomic environments. This variant parameterizes the coupling matrix \textit{via} low-rank factorization $\mathbf{J} = \mathbf{U}\mathbf{U}^\top$, where $\mathbf{U} \in \mathbb{R}^{100 \times 64}$ is a fixed learnable parameter matrix; the low-rank constraint limits the degrees of freedom in $\mathbf{J}$, while the factorization guarantees positive semi-definiteness—both stabilizing training and improving generalization. The rows of $\mathbf{U}$ exhibit recurring latent patterns, suggesting the model captures transferable structural regularities among frequently occurring atomic positions. TV-LowRank, by contrast, displays weaker diagonal signals and less coherent off-diagonal structure, reflecting the terminal head's limited ability to impose global order on softmax-learned features; it dynamically generates an $n\times32$ factor graph through the LowRank hypernetwork (see Appendix Figure A4). FV-Original learns fine-grained relative coupling patterns with small absolute magnitudes, demonstrating that relative interaction structure—rather than raw coupling strength—is the primary driver of performance. Hyper variants produce chaotic, unstructured positive and negative couplings, indicative of overfitting to training noise rather than learning generalizable reaction rules. PairMLP yields no discernible global structure, confirming that per-pair independent computation destroys the global coordinated interactions that define the Ising attention formulation. Across all variants, the degree of ordered global structure in $\mathbf{J}$ correlates positively with multi-center exact-match accuracy, providing direct visual evidence that structured inter-atomic coupling is the source of VIA's empirical gains.

\section{Protein Residue Contact Prediction}

As a supplementary benchmark, we apply VIA to protein residue contact prediction to test whether its coupling mechanism generalizes beyond retrosynthesis. We use the \texttt{contact\_prediction\_binary} dataset from BioMap Research (BioMap, 2024), which contains protein sequences with binary contact labels (C$\alpha$ distance $< 8$\,\AA). Inputs are sequence-only local features; no MSA, no 3D coordinates, and no precomputed structural information are provided. We do not compare against AlphaFold-class or MSA-based systems; the goal is to isolate whether the learned coupling matrix captures cooperative folding information.

\subsection{Results}

Under fair training, FV achieves the best internal BestF1 and P@1 (Table~\ref{tab:protein}). A paired bootstrap over 538 proteins suggests that FV significantly outperforms TV (BestF1 $p=0.012$; P@1 $p<0.001$), winning on $1.75\times$ as many proteins (Appendix A3.5). These are internal ablation results, not SOTA claims. Softmax achieves the lowest BestF1 despite competitive precision, reflecting its peaked distribution that inflates precision on a few predictions at the expense of recall---a critical deficiency for cooperative folding.

\begin{table}[h]
\centering
\caption{Protein contact prediction.}
\label{tab:protein}
\small
\begin{tabular}{lccc}
\toprule
Model & BestF1 & P@1 & P@5 \\
\midrule
Softmax & 0.574 & 0.663 & 0.480 \\
Sigmoid (MLP) & 0.637 & 0.890 & 0.485 \\
VIA-Terminal (TV) & 0.650 & 0.891 & 0.485 \\
\textbf{VIA-Full (FV)} & \textbf{0.6695} & \textbf{0.917} & \textbf{0.4897} \\
\bottomrule
\end{tabular}
\vspace{4pt}
\begin{minipage}{\linewidth}
\footnotesize \textit{Note:} All models trained with identical hyperparameters.
\end{minipage}
\end{table}

\begin{figure}[h]
\centering
\includegraphics[width=\linewidth]{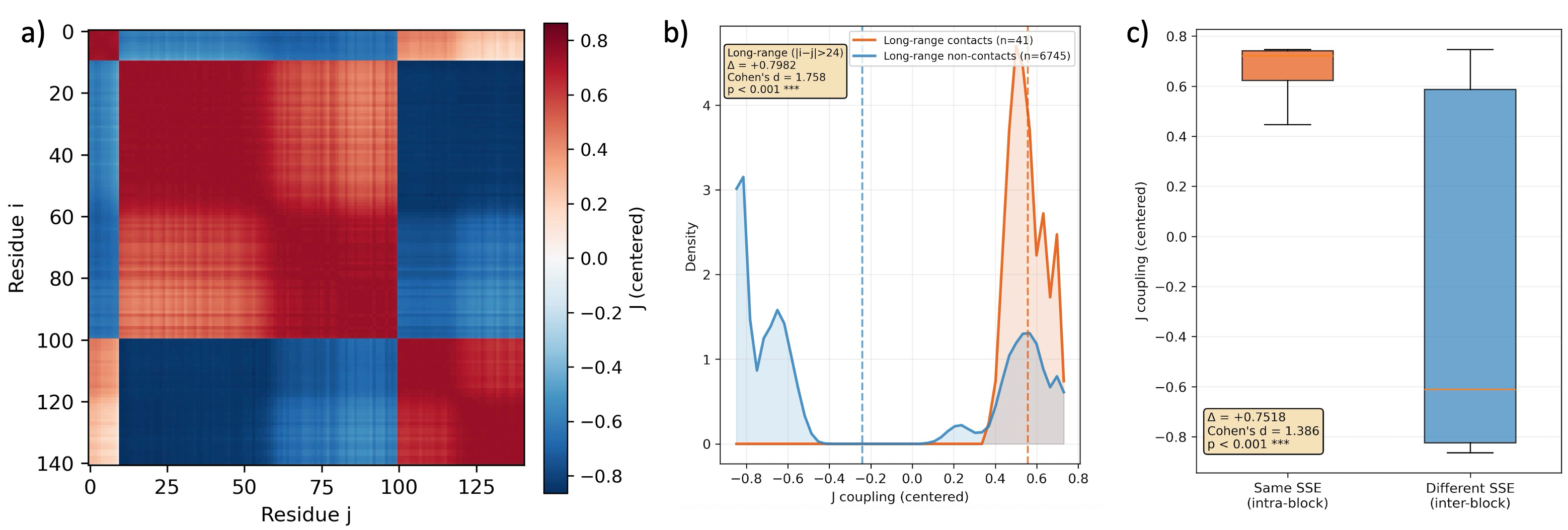}
\caption{Learned coupling matrix $\mathbf{J}$ for a representative protein.}
\label{fig:jmatrix}
\end{figure}

\subsection{What VIA Captures}

The ablation's value lies not in the scores but in what the learned coupling matrix reveals. We visualize $\mathbf{J}$ for a deliberately challenging protein ($L=141$, BestF1$=0.376$). Fig.4a shows the centered $\mathbf{J}$ matrix, which exhibits clear block-diagonal structure aligned with secondary structure elements. Despite the poor per-residue prediction, $\mathbf{J}$ strongly discriminates long-range contacts from non-contacts: Cohen's $d=1.758$ for long-range pairs versus $d=0.752$ for short-range pairs (Fig.4b). VIA does not merely recapitulate local proximity---it specifically captures long-range cooperative folding signals.

Two further diagnostics reinforce this. First, the learned $\mathbf{J}$ has \textbf{effective rank 2.21} (numerical rank 25, top-10 singular values accounting for $99.9\%$ of energy), consistent with the intrinsic two-dimensional physical manifold of folding (hydrophobicity and backbone hydrogen-bond directionality). Second, despite $46.5\%$ negative couplings, the matrix exhibits \textbf{zero geometric frustration}: all four frustrated triadic sign patterns have count zero, while all four non-frustrated patterns are populated, and residues within the same secondary structure element show significantly stronger coupling than those across elements (Fig.4c). In statistical mechanics, zero frustration is the signature of a funnel-shaped energy landscape---no local traps, a unique global minimum. These results show that VIA's learned couplings acquire the same qualitative signatures observed in retrosynthesis---ordered block structure, low frustration, domain-determined effective rank, and strong long-range selectivity---even on a protein the model predicts poorly, indicating that VIA's advantage is structural and physical rather than an artifact of one domain's data statistics.

\begin{table*}[h]
\centering
\caption{Stratification by number of reaction centers $k$ (test set).}
\label{tab:shortcut}
\small
\begin{tabular}{lcccccc}
\toprule
$k$ & $n$ & Slot ceiling & Softmax & PairMLP (equiv.) & LowRank & Original \\
\midrule
1 & 1{,}309 & 91.3\% & 28.6\% & 86.3\% & 97.9\% & 97.7\% \\
2 & 3{,}361 & 34.8\% & 3.2\% & 74.2\% & 94.4\% & 93.4\% \\
3 & 51 & 9.8\% & 0\% & 49.0\% & 64.7\% & 80.4\% \\
$\geq 4$ & 61 & 3.3\% & 0\% & 45.9\% & 82.0\% & 70.5\% \\
\midrule
All & 4{,}782 & 49.5\% & --- & 76.9\% & 94.9\% & 94.1\% \\
Multi ($k\geq 2$) & 3{,}473 & 33.8\% & 3.1\% & 73.3\% & 93.8\% & 92.8\% \\
Shortcut-impossible & 2{,}412 & 0\% & --- & 78.2\% & 90.8\% & 89.3\% \\
\bottomrule
\end{tabular}

\vspace{4pt}
\begin{minipage}{\linewidth}
\footnotesize \textit{Note:} ``Slot ceiling'' is the strongest position-only predictor. ``Shortcut-impossible'' excludes molecules whose true RC set equals the first $k$ slots.
\end{minipage}
\end{table*}

\section{Discussion and Outlook}

Through extensive experiments, quantitative benchmarking, and mechanistic analyses, this work establishes Variational-Ising-Attention (VIA) as a robust, theoretically grounded augment to standard softmax attention for scientific tasks characterized by intrinsic structural interactions. We demonstrate that even a minimal terminal implementation yields measurable gains, whereas full backbone integration unlocks substantially stronger performance by enabling structured couplings to shape the entire feature learning process.

VIA is designed for scientific tasks with intrinsic, data-learnable coupling structure; where couplings are absent or positional, general-purpose attention remains appropriate. Our notion of \emph{tailored} also entails domain-specific customization alongside trade-offs bounding its applicable scenarios, as follows.

\subsection{Atom-order Trade-off}

USPTO-50k products are not canonicalized and remapped by default, and $\mathbf{J}=\mathbf{U}\mathbf{U}^\top$ in VIA is indexed by padded atom positions; Somnath et al.\ (2021) documented that this benchmark contains a position-dependent shortcut unless such preprocessing is performed. To avoid information leakage from SMILES augmentation, we retain the original atom ordering, so the position-indexed coupling is not permutation-equivariant. We quantify the scope of this trade-off below.

Table~\ref{tab:shortcut} shows that a position-only predictor has a hard ceiling of 49.5\% exact match, versus 94.9\% for VIA-Full (LowRank), and that the model retains 90.8\% on the 2{,}412 molecules where no position rule can be correct. The advantage is largest precisely where position priors vanish---multi-center reactions---indicating that the gain reflects chemically meaningful coupling rather than an atom-order shortcut.

\subsection{Mean-Field Iteration Depth}

We characterize the effect of mean-field iteration depth $K$ on both benchmarks. Ranking quality is invariant to $K$, while threshold-calibrated metrics degrade with more iterations---so $K=1$ is both optimal and cheapest. On retrosynthesis, exact-match accuracy is essentially unchanged for $K \in \{1,3,5,7\}$. On protein, using the same FV checkpoint trained with $K=8$, we evaluate $K \in \{1,3,5,7,8\}$ at inference time: ranking metrics are exactly invariant to $K$ (P@1 $=0.917$, P@5 $=0.490$ throughout), because each update applies a monotone softmax over the same coupled field, rescaling probabilities without changing their order; BestF1, however, decreases from $0.6695$ at $K=1$ to $0.6591$ at $K=8$, as accumulated coupling shifts the optimal threshold. Full numerical results are provided in Appendix~A3.6.

This suggests that VIA's advantage is structural rather than computational---the learned $\mathbf{J}$, not the depth of inference, carries the signal. We therefore adopt $K=1$, which is simultaneously optimal for BestF1 and $8\times$ faster, retaining $K$ as an engineering parameter for tasks with more rugged landscapes.

\subsection{Outlook}

While further refinements to VIA's parameterization and broader evaluations across additional scientific benchmarks are underway, the present work already furnishes a complete, empirically validated paradigm for tailored attention design, together with clear principles for constructing structured interaction mechanisms aligned with domain structure.

More broadly, our results suggest that for scientific problems with intrinsic structural coupling, the optimal attention mechanism is not a more efficient general-purpose alternative, but one appropriately tailored to domain-specific structure---an insight for which VIA provides a theoretically grounded and empirically validated instantiation.

\subsection*{AI use statement}

In this work, we used generative AI tools for the following tasks with required disclosure: some content in Figure~1, which serves as an illustrative demonstration figure, was initially generated using OpenAI Codex and then manually edited by the authors. Additionally, we used generative AI tools for the following task with recommended disclosure: portions of the manuscript were polished for grammar, clarity, and style using a large language model. We have reviewed all AI-assisted work. We take responsibility for the final content of this work, including text, claims, or artifacts produced with the aid of generative AI.

\clearpage
\appendix
\thispagestyle{empty}

\section*{Appendix}

\renewcommand{\thefigure}{A\arabic{figure}}
\setcounter{figure}{0}

\renewcommand{\thetable}{A\arabic{table}}
\setcounter{table}{0}

\subsection*{A1. Implementation Details of VIA}

This appendix provides the implementation details of VIA, including the interaction-state updates, coupling-matrix parameterizations, deployment paradigms, and hyperparameter settings. Because TV and FV use different interaction-state parameterizations, their implementations are described separately. To align the terminology with the codebase, we note that for retrosynthesis, TV corresponds to the terminal-layer deployment, while FV corresponds to the backbone deployment.

\subsection*{A1.1 Tensor Shapes through the VIA Pipeline}
\label{app:tensor-shapes}

\begin{figure*}[t]
\centering
\includegraphics[width=0.9\linewidth]{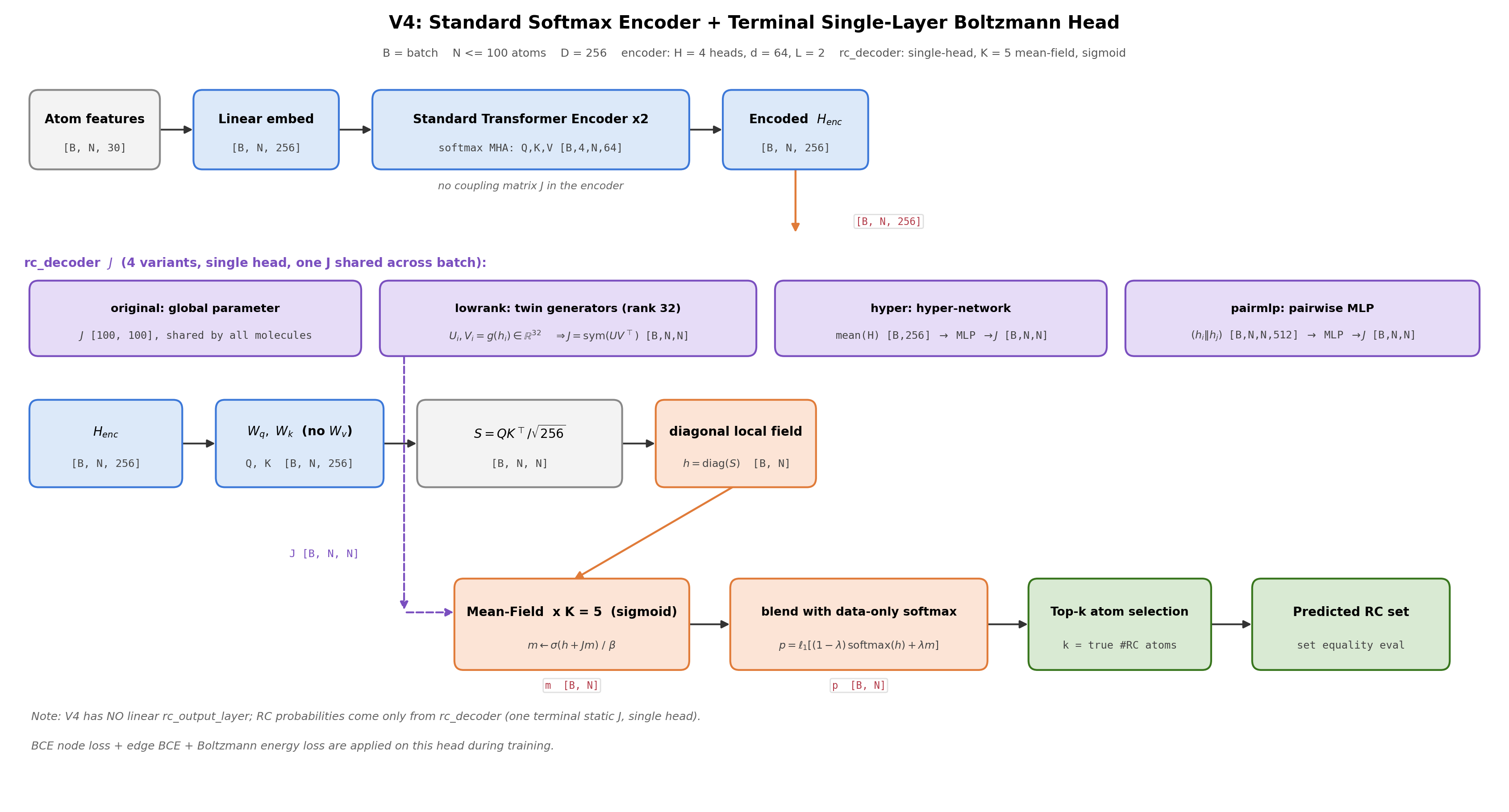}
\caption{Tensor shapes through the VIA forward pass, TV (terminal deployment): the coupling module is applied only at the prediction head; the encoder retains standard self-attention.}
\label{fig:tensor-shapes-tv}
\end{figure*}

\begin{figure*}[t]
\centering
\includegraphics[width=0.9\linewidth]{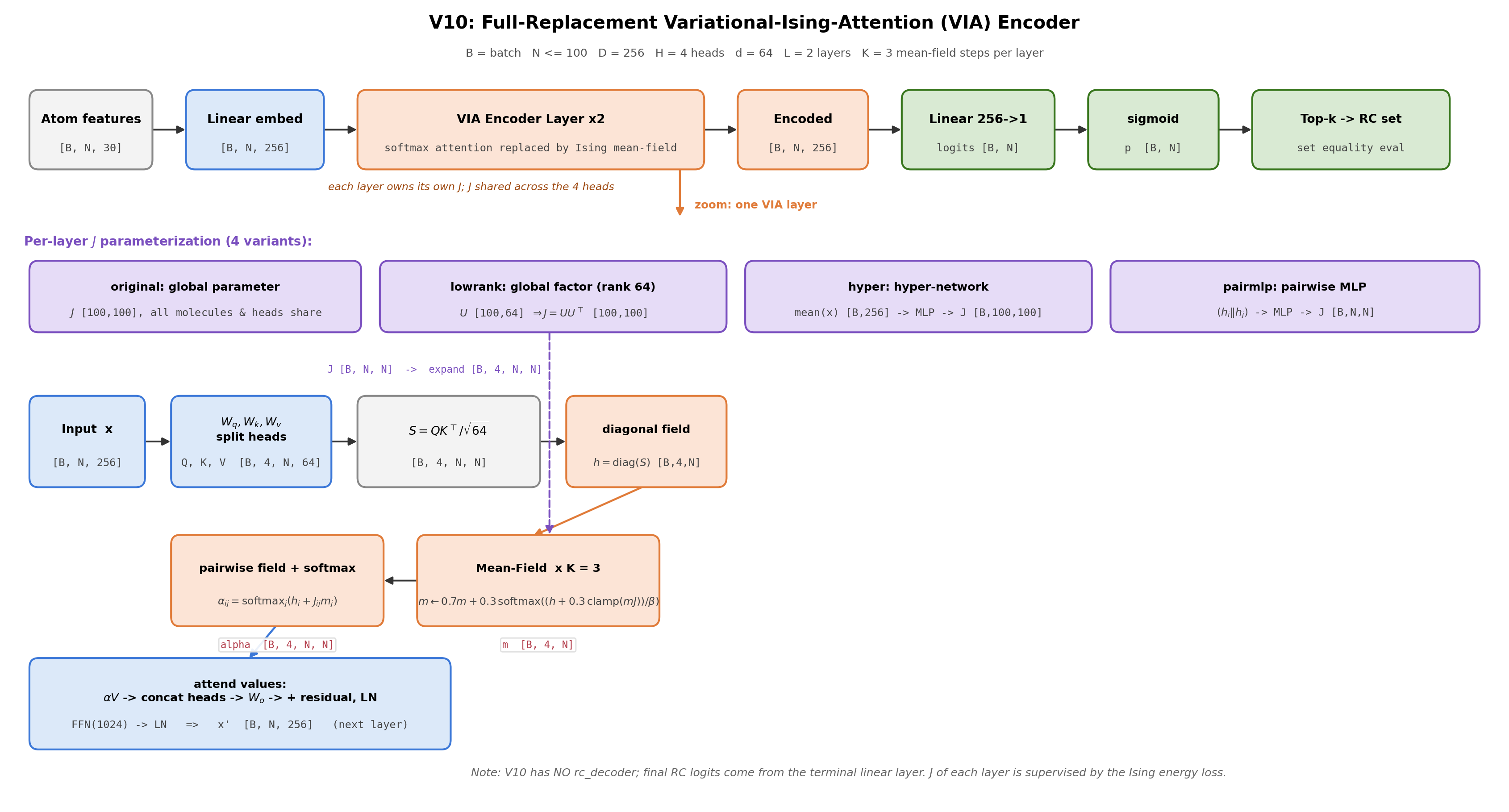}
\caption{Tensor shapes through the VIA forward pass, FV (backbone deployment): VIA replaces self-attention in every encoder layer, so coupling modulates features from the first layer onward.}
\label{fig:tensor-shapes-fv}
\end{figure*}

Figures~\ref{fig:tensor-shapes-tv} and~\ref{fig:tensor-shapes-fv} illustrate the tensor shapes at each stage of the VIA forward pass for both deployment regimes. Let $B$ denote batch size, $H$ the number of attention heads, $N$ the padded sequence length, and $d_k$ the per-head key dimension.

\begin{itemize}
    \item \textbf{Query / Key / Value.} $\mathbf{Q},\mathbf{K},\mathbf{V}\in\mathbb{R}^{B\times H\times N\times d_k}$.
    \item \textbf{Compatibility.} $\mathbf{S}=\mathbf{Q}\mathbf{K}^\top/\sqrt{d_k}\in\mathbb{R}^{B\times H\times N\times N}$.
    \item \textbf{Unary local field.} $h=\mathrm{diag}(\mathbf{S})\in\mathbb{R}^{B\times H\times N}$ (one scalar per position per head).
    \item \textbf{Coupling matrix.} $\mathbf{J}\in\mathbb{R}^{N\times N}$ (shared across heads within a layer; layer-specific).
    \item \textbf{Interaction state.} $m\in\mathbb{R}^{B\times H\times N}$ (simplex-constrained in FV; coordinate-wise in TV).
    \item \textbf{Attention field.} $e_{ij}=h_i+\gamma J_{ij}m_j$, giving $e\in\mathbb{R}^{B\times H\times N\times N}$.
    \item \textbf{Attention weights.} $\alpha=\mathrm{softmax}_j(e/T)\in\mathbb{R}^{B\times H\times N\times N}$.
    \item \textbf{Output.} $\mathbf{Z}=\alpha\mathbf{V}\in\mathbb{R}^{B\times H\times N\times d_k}$.
\end{itemize}

The two regimes differ at two points: (i) the interaction-state update (coordinate-wise sigmoid in TV, simplex-constrained softmax with damping in FV), and (ii) the final aggregation (TV mixes $m$ with the data-only softmax, while FV performs a second coupling-aware softmax over $Jm$).

\subsubsection*{A1.2 Backbone Deployment: V10}

In the backbone deployment, VIA replaces the standard self-attention
operation within the Transformer encoder. The interaction state is
represented as a normalized distribution over sequence positions and
is updated through a softmax-based, damped mean-field-inspired
iteration.

The complete forward computation for a single attention head is
summarized in Algorithm~\ref{alg:via_app}.

\begin{algorithm}[H]
\caption{VIA forward pass for backbone deployment (V10, single head)}
\label{alg:via_app}
\begin{algorithmic}[1]
\REQUIRE
    Query $\mathbf{Q} \in \mathbb{R}^{N \times d_k}$,
    Key $\mathbf{K} \in \mathbb{R}^{N \times d_k}$,
    Value $\mathbf{V} \in \mathbb{R}^{N \times d_v}$,
    coupling $\mathbf{J} \in \mathbb{R}^{N \times N}$,
    temperature $T$,
    coupling scale $\gamma$,
    damping coefficient $\lambda$,
    iterations $K$

\STATE Compute compatibility:
\[
\mathbf{S}
\gets
\mathbf{Q}\mathbf{K}^{\top}/\sqrt{d_k}
\]

\STATE Extract diagonal local fields:
\[
h_j \gets S_{jj}
\]

\STATE Initialize normalized interaction state:
\[
m_j^{(0)}
\gets
\operatorname{softmax}_j(h_j/T)
\]

\FOR{$t=1$ to $K$}

    \STATE Compute pairwise interaction:
    \[
    \delta_j^{(t)}
    \gets
    \sum_{k\neq j}J_{jk}m_k^{(t-1)}
    \]

    \STATE Compute normalized interaction update:
    \[
    \tilde{m}_j^{(t)}
    \gets
    \operatorname{softmax}_j
    \left(
    \frac{h_j+\gamma\delta_j^{(t)}}{T}
    \right)
    \]

    \STATE Apply damping:
    \[
    m_j^{(t)}
    \gets
    \lambda m_j^{(t-1)}
    +(1-\lambda)\tilde{m}_j^{(t)}
    \]

\ENDFOR

\STATE Construct coupling-aware attention field:
\[
\mathrm{field}_{ij}
\gets
h_i+\gamma J_{ij}m_j^{(K)}
\]

\STATE Normalize attention:
\[
\alpha_{ij}
\gets
\operatorname{softmax}_j
\left(
\mathrm{field}_{ij}/T
\right)
\]

\STATE Aggregate values:
\[
\mathbf{Z}
\gets
\boldsymbol{\alpha}\mathbf{V}
\]

\RETURN $\mathbf{Z}$, $\boldsymbol{\alpha}$
\end{algorithmic}
\end{algorithm}

\subsubsection*{A1.3 Terminal Deployment: V4}

In the terminal deployment, VIA is applied only to the
reaction-center decoder while the Transformer encoder retains
standard self-attention. Unlike V10, the V4 interaction state is
updated coordinate-wise with a sigmoid function:
\begin{equation}
    m_j^{(t)}
    =
    \sigma\left(
    \frac{
    h_j+\gamma\sum_{k\neq j}J_{jk}m_k^{(t-1)}
    }{T}
    \right).
\end{equation}

The resulting interaction state is combined with the data-only
attention distribution at the terminal prediction stage:
\begin{equation}
    p
    =
    (1-\lambda)\operatorname{softmax}(h)
    +
    \lambda m.
\end{equation}

Here, $m$ is coordinate-wise rather than simplex-constrained,
allowing multiple positions to be activated simultaneously. This
formulation is therefore used for the multi-label reaction-center
prediction setting, whereas the normalized interaction state in V10
serves as an internal attention-routing representation.

\subsubsection*{A1.4 Coupling Matrix Parameterizations}

We evaluate four parameterizations of $\mathbf{J}$ to examine
different assumptions about relational structure. The full matrix
provides a high-capacity reference, while the low-rank formulation
tests whether pairwise interactions can be represented in a compact
latent space. The hypernetwork formulation allows the coupling
pattern to depend on the input representation, whereas the pairwise
MLP predicts pair-specific interactions from pair representations.

In the backbone deployment (V10), we use the LowRank parameterization
\begin{equation}
    \mathbf{J}=\mathbf{U}\mathbf{U}^{\top},
\end{equation}

with rank $r=64$. This parameterization provides a compact
coupling representation and constrains $\mathbf{J}$ to be positive
semidefinite.

\begin{table*}[t]
\centering
\setlength{\tabcolsep}{4pt}
\footnotesize
\caption{Parameterizations of the coupling matrix $\mathbf{J}$.}
\label{tab:j_param_app}
\begin{tabular}{@{}p{0.08\textwidth}p{0.17\textwidth}p{0.13\textwidth}p{0.09\textwidth}p{0.11\textwidth}p{0.24\textwidth}@{}}
\toprule
Variant & Definition & Source of $\mathbf{J}$ & Input-dependent
& Parameters & Key Property \\
\midrule
Original
& $J_{ij}=\theta_{ij}$
& Learnable parameters
& No
& $N^2$
& Full-rank, unconstrained \\

LowRank
& $\mathbf{J}=\mathbf{U}\mathbf{U}^{\top}$
& Learnable $\mathbf{U}$
& No
& $N\times r$
& Compact, PSD \\

Hyper
& $\mathbf{J}=\mathrm{MLP}(\mathrm{mean}(\mathbf{X}))$
& Hypernetwork
& Yes
& Network weights
& Input-conditioned global coupling \\

PairMLP
& $J_{ij}=\mathrm{MLP}([\mathbf{x}_i;\mathbf{x}_j])$
& Per-pair MLP
& Yes
& Network weights
& Input-conditioned pairwise coupling \\
\bottomrule
\end{tabular}

\end{table*}

\subsubsection*{A1.5 Deployment Paradigms}

We evaluate VIA under two deployment paradigms. Their main
differences are summarized in Table~\ref{tab:deploy_app}. In V4,
the interaction mechanism is introduced only at the terminal
reaction-center decoder, leaving the Transformer encoder unchanged.
In V10, VIA replaces self-attention throughout the Transformer
backbone, allowing interaction-mediated routing to be applied at
each encoder layer.

\begin{table*}[t]
\centering
\setlength{\tabcolsep}{4pt}
\footnotesize
\caption{Deployment paradigms of VIA.}
\label{tab:deploy_app}
\begin{tabular}{@{}lllll@{}}
\toprule
Paradigm & Model & Encoder & Prediction Head & VIA Scope \\
\midrule
Terminal
& V4
& Standard self-attention
& VIA decoder
& Terminal only \\

Backbone
& V10
& VIA
& Sigmoid RC head
& Encoder backbone \\
\bottomrule
\end{tabular}

\end{table*}

\subsubsection*{A1.6 Hyperparameter Settings}

For the VIA interaction updates, we use
$T=1.0$, $\gamma=0.3$, $\lambda=0.7$, and
$K=3$ mean-field-inspired iterations in all experiments.

For the LowRank parameterization, the rank is set to $r=64$ for
sequence lengths up to $512$. The coupling parameters are initialized
from a normal distribution with zero mean and standard deviation
$0.01$, and are trained jointly with the rest of the network.

For backbone deployment, each VIA encoder layer maintains its own
coupling parameterization. The resulting coupling matrix is shared
across attention heads within the same layer, while different encoder
layers use independent coupling parameters.

\subsection*{A2. General Experimental Details of Retrosynthesis RC Prediction}

This appendix provides the detailed atom features, diagnostic definitions, and computational resources omitted from the main text. The coupling matrix parameterizations are described in Appendix~A1, Table~\ref{tab:j_param_app}.

\subsubsection*{A2.1 Atom Features}

All input molecules are parsed by RDKit into 30-dimensional atom features encoding local chemical environments. The full feature list is provided in Table~\ref{tab:atom_features}. All features are clipped to $[-1.0, 1.0]$ after normalization.

\begin{table*}[t]
\centering
\footnotesize
\caption{30-dimensional atom features used as model input.}
\label{tab:atom_features}
\begin{tabular}{@{}clll@{}}
\toprule
Index & Feature Name & Description & Normalization \\
\midrule
0 & Atomic Number & Atomic number of the atom & $/100.0$ \\
1 & Formal Charge & Formal charge on the atom & $/5.0$ \\
2 & Is Aromatic & Whether atom is aromatic & 0/1 binary \\
3 & Total H Count & Number of hydrogens bonded & $/8.0$ \\
4 & Degree & Number of bonded neighbors & $/8.0$ \\
5 & Hybridization SP & One-hot for SP hybridization & 0/1 binary \\
6 & Hybridization SP2 & One-hot for SP2 hybridization & 0/1 binary \\
7 & Hybridization SP3 & One-hot for SP3 hybridization & 0/1 binary \\
8 & Mass & Atomic mass & $/200.0$ \\
9 & Is In Ring & Whether atom belongs to a ring & 0/1 binary \\
10 & Ring Size & Size of smallest ring containing atom & $\min(\text{size},12)/12.0$ \\
11 & Implicit Valence & Implicit valence of the atom & $/8.0$ \\
12 & Explicit Valence & Explicit valence of the atom & $/8.0$ \\
13--18 & Neighbor Atomic Numbers & Atomic numbers of up to 6 neighbors & each $/100.0$ \\
19 & Gasteiger Charge & Gasteiger partial charge & $/2.0$ \\
20 & Avg Bond Order & Average bond order to neighbors & raw value \\
21 & Max Bond Order & Maximum bond order to neighbors & raw value \\
22 & Min Bond Order & Minimum bond order to neighbors & raw value \\
23 & Valence Difference & Implicit valence minus explicit valence & $/8.0$ \\
24 & Is Saturated & Whether atom is fully saturated & 0/1 binary \\
25 & Avg Neighbor Atomic Num & Average atomic number of neighbors & $/100.0$ \\
26 & Max Neighbor Atomic Num & Maximum atomic number of neighbors & $/100.0$ \\
27 & Heavy Atom Ratio & Proportion of heavy-atom neighbors & raw ratio \\
28 & Hetero Atom Ratio & Proportion of hetero-atom neighbors & raw ratio \\
29 & Aromatic Neighbor Ratio & Proportion of aromatic neighbors & raw ratio \\
\bottomrule
\end{tabular}

\end{table*}

These features encode purely \textit{local} chemical environments and do not include any pre-computed global structural information such as 3D coordinates or explicit atom mapping indices. Labels (reaction centers) are derived separately by comparing atom mappings between product and reactant molecules, ensuring no information leakage from labels to input features.

\subsubsection*{A2.2 Evaluation Metrics Details.}
For atom-level binary classification, let TP, FP, FN denote true positives, false positives, and false negatives at threshold 0.5. Then:

\[
\begin{aligned}
\text{Precision} &= \frac{\text{TP}}{\text{TP}+\text{FP}}, \\
\text{Recall} &= \frac{\text{TP}}{\text{TP}+\text{FN}}, \\
\text{F1} &= 2 \cdot \frac{\text{Precision} \cdot \text{Recall}}{\text{Precision} + \text{Recall}}.
\end{aligned}
\]
In our empirical setting, top-$k$ exact-match and atom-level F1 are positively correlated: better atom-level ranking directly increases the probability that all $k$ reactive atoms occupy the top-$k$ positions, and we observe that F1 gains consistently translate to exact-match improvements across models.

\subsubsection*{A2.3 Training Dynamics.}

Figure A3 compares the learning dynamics of V4 (terminal) and V10 (backbone) deployment across all four coupling parameterizations, alongside their respective softmax attention baselines. Consistent patterns emerge across both architectural configurations.

\begin{figure*}[!t]
\centering
\includegraphics[width=0.9\textwidth]{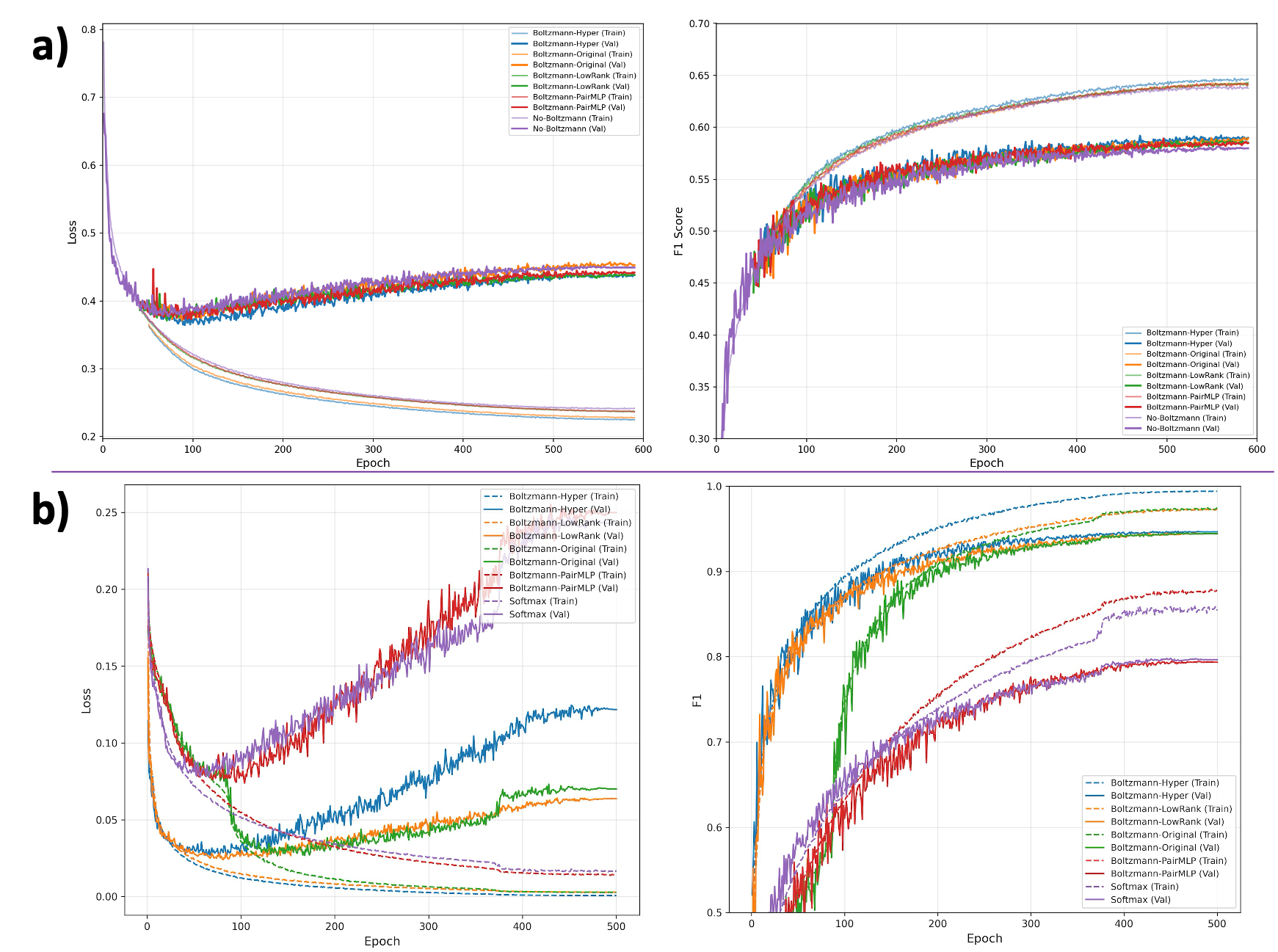}
\caption{Training Dynamics}
\label{f2p}
\end{figure*}

For terminal deployment (Figure A3a), all VIA variants converge within the first 100 epochs and outperform the softmax baseline in both loss decay and atom-level F1. Backbone deployment (Figure A3b) accelerates convergence and elevates the performance ceiling across all parameterizations except PairMLP. V10-LowRank, V10-Hyper, and V10-Original all exceed 0.9 atom-level F1 within 200 epochs and continue to improve to above 0.95 by epoch 500; their validation F1 also reaches around 0.95 by the end of training.

\subsubsection*{A2.4 Diagnostic Definitions}

\paragraph{Geometric Frustration Index.}
For a coupling matrix $\mathbf{J} \in \mathbb{R}^{N \times N}$ over real atoms:
\[
f = \frac{1}{\binom{N}{3}} \sum_{i<j<k} |J_{ij} J_{jk} J_{ki}|,
\]
where $f \approx 0$ indicates a smooth, frustration-free landscape, and $f \to 1$ indicates maximal triadic conflict.

\paragraph{Effective Rank.}
Let $\sigma_k$ be the singular values of $\mathbf{J}$, and $\tilde{\sigma}_k = \sigma_k / \sum_{k'} \sigma_{k'}$. Then:
\[
r_{\text{eff}} = \exp\left( -\sum_{k=1}^{N} \tilde{\sigma}_k \log \tilde{\sigma}_k \right).
\]
This measures the number of independent modes represented by $\mathbf{J}$.

\subsubsection*{A2.5 Computational Resources}

All experiments were conducted on a single NVIDIA A100 (40GB) GPU. Training time for V10-LowRank is approximately 6 hours for 500 epochs, which is comparable to the standard softmax attention baseline. V10-Original takes approximately 7 hours, while V10-Hyper requires approximately 8 hours for the same number of epochs. V10-PairMLP is notably slower, taking roughly 11 hours to complete 500 epochs.

\subsubsection*{A2.6 Supplementary Mechanism Results}

\paragraph*{Effective Rank Stratification (All Variants)}

Table~\ref{app:tab:rank} stratifies test accuracy by effective rank. Original, Hyper, and PairMLP exhibit an inverted-U peaking at $[8,12)$, while LowRank monotonically decreases, highlighting its built-in bottleneck regularization. Interestingly, LowRank exhibits a positive rather than negative association between geometric frustration and prediction accuracy across the observed strata. This result indicates that geometric frustration alone does not explain the predictive advantage of VIA. We therefore interpret frustration and effective rank as descriptive diagnostics of the learned coupling structure rather than causal explanations of performance.

\begin{table}[htbp]
\centering
\setlength{\tabcolsep}{3pt}
\footnotesize
\caption{Test accuracy (\%) stratified by effective rank. -- indicates no samples.}
\label{app:tab:rank}
\begin{tabular}{lcccc}
\toprule
\textbf{Rank} & \textbf{Original} & \textbf{LowRank} & \textbf{Hyper} & \textbf{PairMLP} \\
\midrule
$[1, 3)$ & -- & -- & -- & 0.00 \\
$[3, 5)$ & -- & 97.22 & 100.00$^\dagger$ & 43.48 \\
$[5, 8)$ & 94.44 & 95.65 & 95.34 & 81.03 \\
$[8, 12)$ & 96.49 & 86.70 & 96.10 & 83.63 \\
$[12, 20)$ & 96.17 & -- & 94.32 & 75.53 \\
$[20, 100)$ & 93.75 & -- & 79.05 & 72.32 \\
\bottomrule
\end{tabular}
\par\medskip
\footnotesize $^\dagger$ $n=4$; excluded from trend.

\end{table}

\paragraph*{Geometric Frustration Stratification (LowRank)}

Table~\ref{app:tab:frust} details the counter-intuitive positive correlation for LowRank.

\begin{table}[htbp]
\centering
\setlength{\tabcolsep}{4pt}
\footnotesize
\caption{LowRank accuracy by frustration interval.}
\label{app:tab:frust}
\begin{tabular}{lcc}
\toprule
\textbf{Frustration interval} & \textbf{Accuracy (\%)} & \textbf{N} \\
\midrule
$[0.01, 0.02)$ & 88.89 & 18 \\
$[0.02, 0.05)$ & 93.14 & 102 \\
$[0.05, 0.10)$ & 95.87 & 218 \\
$[0.10, 0.20)$ & 97.05 & 102 \\
$[0.20, 1.00)$ & 97.98 & 99 \\
\bottomrule
\end{tabular}

\end{table}

\subsubsection*{A2.7 Energy Perturbation Robustness}

We evaluate structural integrity \textit{via} (i) Gaussian noise (Table~\ref{app:tab:noise}), (ii) sign flips (Table~\ref{app:tab:flip}), and (iii) low-rank truncation (Table~\ref{app:tab:trunc}). LowRank achieves the best balance of accuracy and robustness.

\begin{table}[htbp]
\centering
\setlength{\tabcolsep}{3pt}
\footnotesize
\caption{Mean $|\Delta E|$ under additive Gaussian noise.}
\label{app:tab:noise}
\begin{tabular}{lcccc}
\toprule
$\sigma$ & Original & LowRank & Hyper & PairMLP \\
\midrule
0.01 & 0.0041 & 0.0044 & 0.0287 & 0.0065 \\
0.05 & 0.0199 & 0.0218 & 0.1436 & 0.0332 \\
0.10 & 0.0394 & 0.0450 & 0.2883 & 0.0691 \\
0.20 & 0.0804 & 0.0902 & 0.5759 & 0.1425 \\
\bottomrule
\end{tabular}

\end{table}

\begin{table}[htbp]
\centering
\setlength{\tabcolsep}{3pt}
\footnotesize
\caption{Mean $|\Delta E|$ under random sign flips.}
\label{app:tab:flip}
\begin{tabular}{lcccc}
\toprule
\textbf{Flip prop.} & Original & LowRank & Hyper & PairMLP \\
\midrule
0.05 & 0.0165 & 0.1878 & 0.9552 & 0.3966 \\
0.10 & 0.0335 & 0.3394 & 1.9449 & 0.7660 \\
0.20 & 0.0673 & 0.6609 & 3.9364 & 1.5312 \\
0.30 & 0.1010 & 0.9701 & 5.8776 & 2.2795 \\
\bottomrule
\end{tabular}

\end{table}

\begin{table}[htbp]
\centering
\setlength{\tabcolsep}{3pt}
\footnotesize
\caption{Mean $|\Delta E|$ under low-rank truncation (fraction retained).}
\label{app:tab:trunc}
\begin{tabular}{lcccc}
\toprule
\textbf{Fraction} & Original & LowRank & Hyper & PairMLP \\
\midrule
0.80 & 0.0002 & 0.0000 & 0.0119 & 0.0008 \\
0.40 & 0.0005 & 0.0009 & 0.1694 & 0.0265 \\
0.20 & 0.0010 & 0.0097 & 0.2117 & 0.0801 \\
0.10 & 0.0018 & 0.1176 & 0.1722 & 0.3243 \\
\bottomrule
\end{tabular}

\end{table}

\subsection*{A2.8 Reaction-Class and Molecular-Size Stratification}
\label{app:reaction-class}

To verify that VIA's advantage is not concentrated in a particular reaction class, we stratify test exact-match accuracy by the ten reaction categories used in GraphRetro~ (Somnath et al., 2021). The category labels are recovered by matching our test reactions against the GraphRetro release via (id, mapped product SMILES); all 4{,}782 test reactions are matched with 100\% coverage. Table~\ref{tab:reaction-class} reports results for the three representative models.

\begin{table*}[htbp]
\centering
\setlength{\tabcolsep}{2.5pt}
\footnotesize
\caption{Test exact-match accuracy (\%) stratified by reaction class (oracle-$k$ set matching).}
\label{tab:reaction-class}
\begin{tabular}{cp{0.33\columnwidth}ccc}
\toprule
\textbf{\#} & \textbf{Reaction class} & \textbf{Softmax} & \textbf{LowRank} & \textbf{PairMLP} \\
\midrule
1 & Heteroatom alkylation & 2.2 & 94.8 & 72.1 \\
2 & Acylation & 0.0 & 92.2 & 83.6 \\
3 & C--C bond formation & 1.8 & 92.7 & 53.8 \\
4 & Heterocycle formation & 0.0 & 79.1 & 44.0 \\
5 & Protection & 0.0 & 97.1 & 94.1 \\
6 & Deprotection & 16.0 & 99.8 & 92.5 \\
7 & Reduction & 50.3 & 99.3 & 85.2 \\
8 & Oxidation & 0.0 & 100.0 & 96.3 \\
9 & Functional group interconversion & 3.8 & 96.7 & 64.1 \\
10 & Functional group addition & 0.0 & 95.7 & 95.7 \\
\bottomrule
\end{tabular}
\end{table*}

Three observations follow. First, Softmax collapses on most classes (0--4\%), except on reduction (50.3\%) and deprotection (16.0\%), where breaking is largely localized. Second, the LowRank variant (FV) is uniformly strong (79.1--100\%), with the largest gains on deprotection ($+83.8$~pt), reduction ($+49.0$~pt), and oxidation ($+100$~pt), i.e., precisely the classes where multiple bond-breaking sites must be coordinated. Third, the equivariant PairMLP variant is markedly weaker on C--C bond formation (53.8\%) and heterocycle formation (44.0\%), which we attribute to its inability to bind couplings to specific atom positions; this is consistent with the position-prior analysis in Section~5.1.

\paragraph{Molecular size.} Accuracy decreases monotonically with molecule size for all models, but the relative ordering is preserved: on the largest molecules ($\geq 41$ atoms), FV retains 89.8\% exact match while PairMLP drops to 62.3\%. This is consistent with the mechanistic finding that VIA's gain grows with reactive complexity.

\paragraph{Directional error patterns.} Across the full test set, the number of cases where Softmax succeeds and FV fails is only 5, whereas the reverse (FV succeeds, Softmax fails) is 4{,}188---a ${\sim}840{:}1$ directional asymmetry. This extreme asymmetry is difficult to explain by class imbalance or random error, and reinforces the conclusion that FV captures structure that standard attention systematically misses.

\subsubsection*{A2.9 Supplementary Visualization}

\begin{figure*}[t]
\centering
\includegraphics[width=0.75\textwidth]{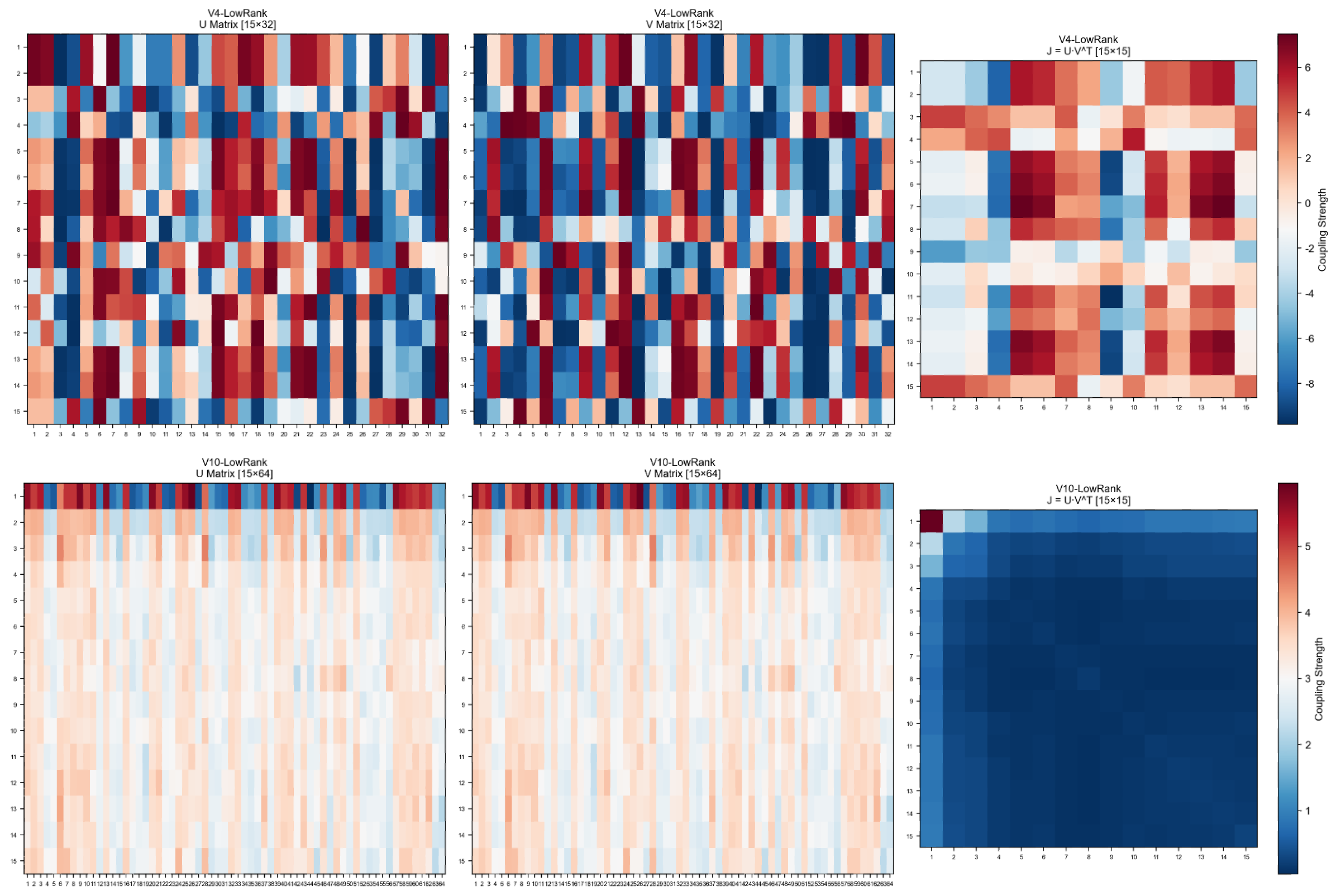}
\caption{Row patterns of the learned factor matrix $\mathbf{U}$ for (top) V4-LowRank and (bottom) V10-LowRank. Rows are clustered by atomic type; chemically similar atoms exhibit correlated latent signatures, with V10-LowRank showing more pronounced block structure.}
\label{app:fig:U}
\end{figure*}

Figure~\ref{app:fig:U} visualizes row patterns of $\mathbf{U}$ for V4 \&V10-LowRank. Rows with similar atomic environments exhibit correlated latent patterns, suggesting that the learned factor matrix captures recurring structural regularities, directly inducing the block-diagonal couplings in the main text.

\subsubsection*{A2.10 Mean-Field Iteration Steps}

We evaluate the effect of mean-field iteration steps \(K \in \{1, 3, 5, 7\}\) on all four V10 backbone variants. Table~\ref{app:tab:mf_steps} reports test-set exact-match accuracy (\%) across reaction center counts. Across all variants and all \(K\), the performance remains essentially unchanged——variations are within \(0.15\%\) for PairMLP and at most \(1.96\%\) for LowRank's 3-RC subset (\(66.67\%\) at \(K=1\) vs. \(64.71\%\) at \(K=3\)). This suggests that the learned \(\mathbf{J}\) matrix is sufficiently well-conditioned that mean-field inference reaches its fixed point within a single iteration. We therefore adopt \(K=1\) for all main results, while retaining \(K\) as a tunable parameter for tasks with more rugged energy landscapes.

\begingroup
\setlength{\tabcolsep}{3pt} 
\scriptsize                    
\begin{table}[htbp]
\centering
\caption{Test exact-match accuracy (\%) for V10 backbone variants across mean-field iteration steps \(K\).}
\label{app:tab:mf_steps}
\resizebox{\columnwidth}{!}{%
\begin{tabular}{lccccccc}
\toprule
\textbf{Model} & \(K\) & \textbf{Overall} & \textbf{1RC} & \textbf{2RC} & \textbf{3RC} & \textbf{4RC} & \textbf{5+} \\
\midrule
Hyper & 1 & 93.35 & 97.40 & 92.68 & 66.67 & 66.67 & 61.54 \\
Hyper & 3 & 93.39 & 97.40 & 92.74 & 66.67 & 66.67 & 61.54 \\
Hyper & 5 & 93.43 & 97.40 & 92.80 & 66.67 & 66.67 & 61.54 \\
Hyper & 7 & 93.43 & 97.40 & 92.80 & 66.67 & 66.67 & 61.54 \\
\midrule
LowRank & 1 & 94.92 & 97.94 & 94.41 & 66.67 & 77.08 & 100.00 \\
LowRank & 3 & 94.90 & 97.94 & 94.41 & 64.71 & 77.08 & 100.00 \\
LowRank & 5 & 94.90 & 97.94 & 94.41 & 64.71 & 77.08 & 100.00 \\
LowRank & 7 & 94.90 & 97.94 & 94.41 & 64.71 & 77.08 & 100.00 \\
\midrule
Original & 1 & 94.14 & 97.78 & 93.37 & 80.39 & 70.83 & 69.23 \\
Original & 3 & 94.14 & 97.78 & 93.37 & 80.39 & 70.83 & 69.23 \\
Original & 5 & 94.14 & 97.78 & 93.37 & 80.39 & 70.83 & 69.23 \\
Original & 7 & 94.14 & 97.78 & 93.37 & 80.39 & 70.83 & 69.23 \\
\midrule
PairMLP & 1 & 76.91 & 86.33 & 74.20 & 50.98 & 45.83 & 46.15 \\
PairMLP & 3 & 76.96 & 86.33 & 74.26 & 50.98 & 45.83 & 46.15 \\
PairMLP & 5 & 77.10 & 86.40 & 74.44 & 50.98 & 45.83 & 46.15 \\
PairMLP & 7 & 77.06 & 86.40 & 74.38 & 50.98 & 45.83 & 46.15 \\
\bottomrule
\end{tabular}%
}

\end{table}
\endgroup

\subsubsection*{A2.11 Gibbs Sampling Ablation Results}

As noted in the main text, we additionally evaluated an alternative inference paradigm for the Ising attention formulation: Boltzmann attention with Gibbs sampling. Specifically, instead of variational mean-field inference, we draw Monte Carlo samples from the Ising energy distribution using sequential Gibbs updates and approximate the attention weights via empirical averages over collected samples. This approach, while conceptually straightforward, suffers from two critical drawbacks: (i) the Markov chain requires hundreds of iterations to mix, especially for molecules with multiple reaction centers, leading to prohibitively slow training (approximately 5--8$\times$ slower per epoch than mean-field VIA); and (ii) the stochastic nature of Gibbs sampling introduces high variance in gradient estimates, causing unstable optimization and inferior final performance.

Tables A10 report the exact-match accuracy on the validation and test sets, respectively, stratified by reaction center count, for all four coupling parameterizations under Gibbs sampling. Across both splits and all variants, performance is substantially worse than the corresponding mean-field VIA counterparts reported in Figure 2d of the main text. The best-performing Gibbs variant, PairMLP, achieves only $73.7\%$ overall validation accuracy and $73.0\%$ on the test set, far below the mean-field V4 models. Notably, the Original parameterization collapses almost entirely under Gibbs sampling (validation $1.3\%$, test $2.1\%$), confirming that the unconstrained full-rank $\mathbf{J}$ matrix produces a rugged energy landscape that frustrates MCMC exploration. These results substantiate our choice of variational mean-field inference as the primary inference engine for VIA.

\begingroup
\setlength{\tabcolsep}{3pt}  
\scriptsize      
\begin{table}[t]
\centering
\caption{Gibbs sampling: exact-match accuracy (\%) by split and reaction center count.}
\label{tab:gibbs}
\resizebox{\columnwidth}{!}{%
\begin{tabular}{lccccccc}
\toprule
\textbf{Split} & \textbf{Model} & \textbf{1RC} & \textbf{2RC} & \textbf{3RC} & \textbf{4RC} & \textbf{5+} & \textbf{Overall} \\
\midrule
\multirow{4}{*}{Val}
 & PairMLP  & 74.8  & 74.2  & 46.6  & 41.9  & 50.0  & 73.7 \\
 & LowRank  & 68.7  & 54.6  & 31.0  & 29.0  & 20.0  & 57.8 \\
 & Hyper    & 68.5  & 39.4  & 12.1  &  0.0  & 25.0  & 46.6 \\
 & Original &  1.5  &  1.2  &  8.6  &  0.0  &  0.0  &  1.3 \\
\midrule
\multirow{4}{*}{Test}
 & PairMLP  & 73.2  & 74.0  & 45.1  & 39.6  & 30.8  & 73.0 \\
 & LowRank  & 67.7  & 53.5  & 17.6  & 20.8  &  7.7  & 56.5 \\
 & Hyper    & 67.1  & 40.9  & 11.8  &  4.2  & 15.4  & 47.3 \\
 & Original &  3.7  &  1.5  &  7.8  &  0.0  &  0.0  &  2.1 \\
\bottomrule
\end{tabular}%
}

\end{table}
\endgroup

The performance degradation under Gibbs sampling is particularly striking. For 3-RC and 4-RC cases, which require coordinated interaction modeling across multiple reactive sites, the stochastic nature of Gibbs sampling fails to reliably explore the joint configuration space, leading to near-zero accuracy for Hyper and Original on 4-RC validation samples ($0.0\%$). These results provide evidence that the deterministic mean-field approximation is better suited for stable end-to-end optimization in the current setting.

\subsection*{A3 Protein Benchmark Details}
\label{app:protein}

This appendix provides supplementary details for the protein residue contact prediction benchmark, including architecture configuration, training setup, statistical validation, and full mechanistic analysis.

\subsubsection*{A3.1 Architecture and Capacity Configuration}

Protein sequences are substantially longer than retrosynthesis molecules ($N_{\max}=512$ vs.\ 100), requiring a separately tuned capacity configuration. VIA's coupling mechanism is identical across both tasks; only capacity is reallocated per domain.

\begin{table}[htbp]
\centering
\setlength{\tabcolsep}{4pt}
\footnotesize
\caption{Cross-task capacity configuration.}
\label{tab:capacity}
\begin{tabular}{p{0.28\columnwidth}p{0.28\columnwidth}p{0.28\columnwidth}}
\toprule
Config & Retrosynthesis & Protein \\
\midrule
$N_{\max}$ & 100 & 512 \\
Layers & 2 & 6 \\
Heads & 8 & 4 \\
Hidden & 512 & 256 \\
FFN & 2048 & 1024 \\
VIA rank $r$ & 64 (FV), 32 (TV) & 64 (FV), 32 (TV) \\
Training epochs & 500 & 300 \\
Mean-field $K$ (training) & 1 & 8 \\
\bottomrule
\end{tabular}
\end{table}

\paragraph{Rationale.} Hidden dimension and heads are reduced (512$\to$256, 8$\to$4) because attention cost scales as $O(N_{\max}^2)$ and per-layer activation memory as $O(N_{\max}\times d_{\text{model}})$. Layers are increased (2$\to$6) not because of sequence length but because protein contact prediction must recover long-range cooperative relationships from purely local sequence features; deeper networks provide the hierarchical abstraction required to propagate global constraints.

\subsubsection*{A3.2 Data and Training}

We use the \texttt{contact\_prediction\_binary} dataset from BioMap Research, containing protein sequences with binary contact labels (C$\alpha$ distance $< 8$\,\AA). Inputs are sequence-only local features; no MSA, no 3D coordinates, and no precomputed structural information are provided.

All protein models train for 300 epochs with learning rate $1\times10^{-4}$ and weight decay 0.01. VIA models use $\mathbf{J}=\mathbf{U}\mathbf{U}^\top$, $T=1.0$, $\gamma=0.3$, $\lambda=0.7$. Training uses $K=8$; inference uses $K=1$. Full FV training takes approximately 12--16 hours on a single NVIDIA A100 (40GB).

\subsubsection*{A3.3 Fair Hyperparameter Ablation}

All four models are trained with identical hyperparameters. Table~\ref{tab:fair300} reports TEST-set results.

\begin{table}[htbp]
\centering
\setlength{\tabcolsep}{2.5pt}
\footnotesize
\caption{Fair 300-epoch TEST results (identical hyperparameters across models). Best in \textbf{bold}.}
\label{tab:fair300}
\begin{tabular}{lccccc}
\toprule
Model & BestF1 & P@1 & P@5 & P@L & Loss \\
\midrule
Standard & 0.637 & 0.890 & 0.485 & 0.185 & 0.329 \\
VIA (pairwise) & 0.650 & 0.891 & 0.485 & 0.186 & 0.320 \\
\textbf{FV (Full VIA)} & \textbf{0.659} & \textbf{0.917} & \textbf{0.490} & \textbf{0.189} & 0.333 \\
Softmax & 0.574 & 0.663 & 0.480 & 0.382 & 2.229 \\
\bottomrule
\end{tabular}
\end{table}

\begin{figure*}[t]
\centering
\includegraphics[width=0.85\textwidth]{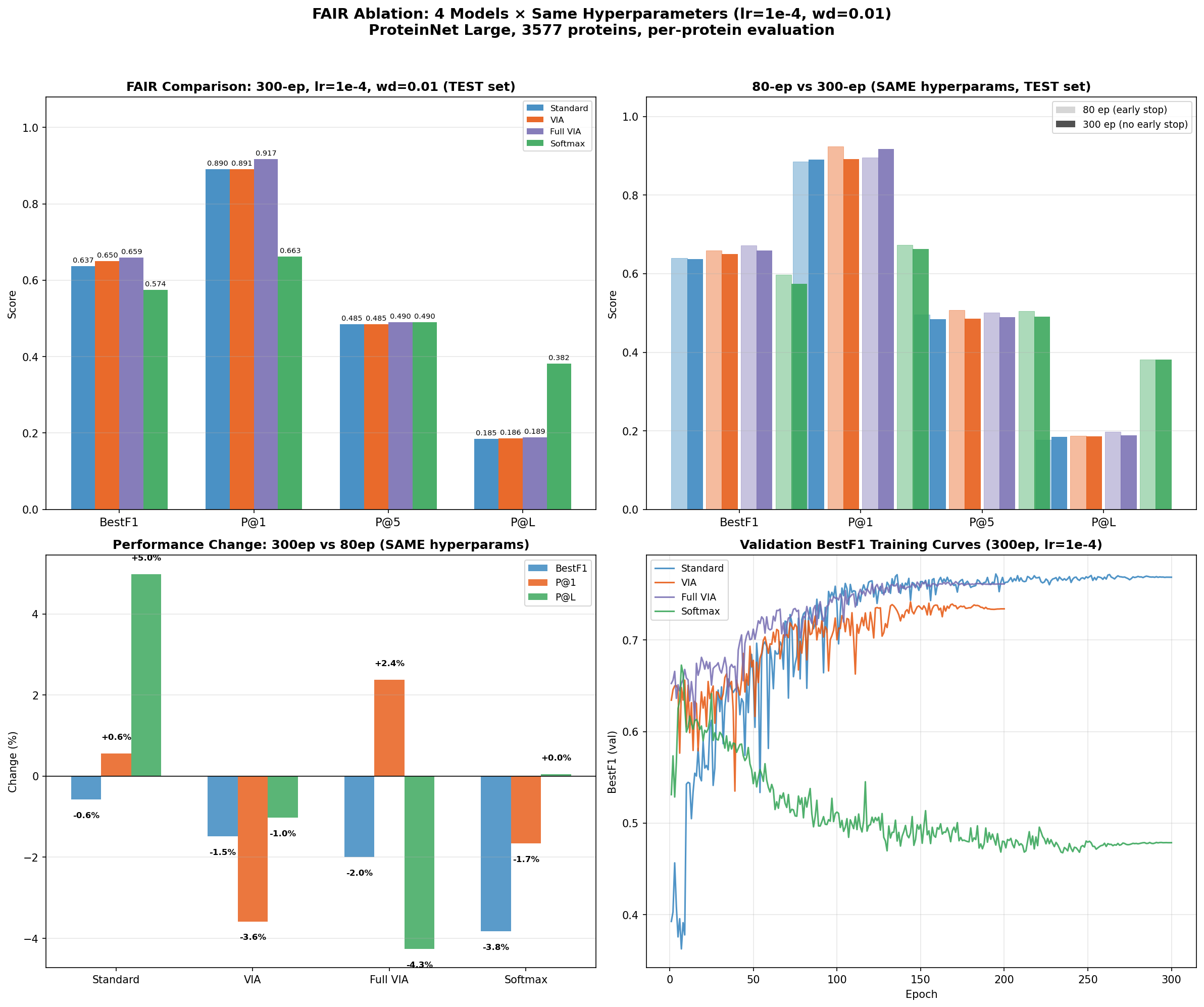}
\caption{Fair 300-epoch comparison across four models. FV achieves the highest BestF1, P@1, P@5, and P@L among BCE-style models; Softmax attains the highest P@L but the lowest BestF1.}
\label{fig:fair300}
\end{figure*}

FV achieves the best BestF1, P@1, P@5, and P@L among BCE-style models. Softmax attains the highest P@L but the lowest BestF1, consistent with its peaked distribution inflating precision on a few predictions at the expense of recall. Figure~\ref{fig:fair300} visualizes this comparison.

\subsubsection*{A3.4 Training-Length Effect: 80ep vs.\ 300ep}

\begin{table}[htbp]
\centering
\setlength{\tabcolsep}{4pt}
\footnotesize
\caption{80ep vs.\ 300ep under identical hyperparameters.}
\label{tab:80v300}
\begin{tabular}{llccc}
\toprule
Model & Metric & 80ep & 300ep & Change \\
\midrule
Standard & BestF1 & 0.640 & 0.637 & $-0.6\%$ \\
Standard & P@L & 0.176 & 0.185 & $+5.0\%$ \\
VIA & BestF1 & 0.659 & 0.650 & $-1.5\%$ \\
VIA & P@1 & 0.925 & 0.891 & $-3.6\%$ \\
FV & BestF1 & 0.673 & 0.659 & $-2.0\%$ \\
FV & P@1 & 0.896 & 0.917 & $\mathbf{+2.4\%}$ \\
Softmax & BestF1 & 0.597 & 0.574 & $-3.8\%$ \\
\bottomrule
\end{tabular}
\end{table}

\begin{figure*}[t]
\centering
\includegraphics[width=0.85\linewidth]{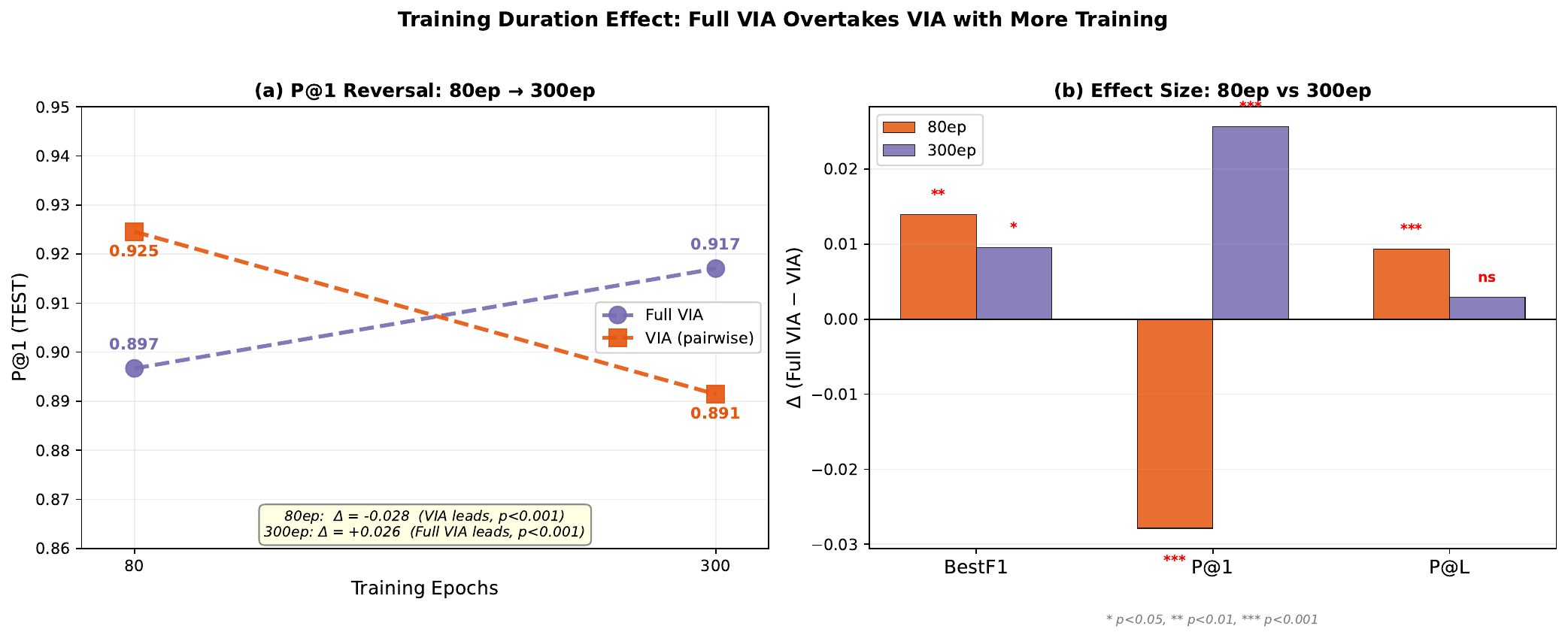}
\caption{Training-duration effect on P@1. At 80 epochs, VIA (pairwise) leads; at 300 epochs, FV reverses the ordering. Dashed lines connect the two measured checkpoints without assuming a linear path.}
\label{fig:p1-crossover}
\end{figure*}

Most changes are within $\pm5\%$, confirming that 80 epochs with early stopping is near convergence. The key exception is FV's P@1, which \emph{improves} with extended training. Figure~\ref{fig:p1-crossover} visualizes this crossover.

\subsubsection*{A3.5 Paired Bootstrap Statistics}

\begin{table}[htbp]
\centering
\setlength{\tabcolsep}{3pt}
\footnotesize
\caption{Paired bootstrap (10{,}000 resamples) over 538 test proteins (300ep, fair hyperparameters).}
\label{tab:bootstrap-full}
\resizebox{\columnwidth}{!}{%
\begin{tabular}{lcccc}
\toprule
Metric & $\Delta$ (FV $-$ VIA) & 95\% CI & $p$-value & Cohen's $d$ \\
\midrule
BestF1 & $+0.0095$ & $[+0.001, +0.018]$ & $0.0118$ & $0.097$ \\
P@1 & $+0.0256$ & $[+0.011, +0.041]$ & $0.0003$ & $0.145$ \\
P@5 & $+0.0045$ & $[-0.002, +0.012]$ & $0.104$ & $0.053$ \\
P@L & $+0.0029$ & $[-0.001, +0.007]$ & $0.058$ & $0.068$ \\
\bottomrule
\end{tabular}%
}
\end{table}

\paragraph{Per-protein win/loss/tie (BestF1, $n=538$).} FV wins 221 (41.1\%), VIA wins 126 (23.4\%), ties 191 (35.5\%); win ratio $1.75{:}1$.

\begin{table}[htbp]
\centering
\setlength{\tabcolsep}{4pt}
\footnotesize
\caption{Full pairwise comparison matrix (BestF1 and P@1, 300ep fair).}
\label{tab:pairwise}
\begin{tabular}{llcc}
\toprule
Metric & Comparison & $\Delta$ & $p$-value \\
\midrule
BestF1 & FV vs.\ Standard & $+0.023$ & $<0.0001$ \\
BestF1 & VIA vs.\ Standard & $+0.013$ & $0.002$ \\
BestF1 & FV vs.\ VIA & $+0.010$ & $0.012$ \\
BestF1 & FV vs.\ Softmax & $+0.085$ & $<0.0001$ \\
P@1 & FV vs.\ Standard & $+0.027$ & $0.0004$ \\
P@1 & VIA vs.\ Standard & $+0.001$ & $0.439$ \\
P@1 & FV vs.\ VIA & $+0.026$ & $0.0003$ \\
P@1 & FV vs.\ Softmax & $+0.254$ & $<0.0001$ \\
\bottomrule
\end{tabular}
\end{table}

\begin{figure*}[t]
\centering
\includegraphics[width=\linewidth]{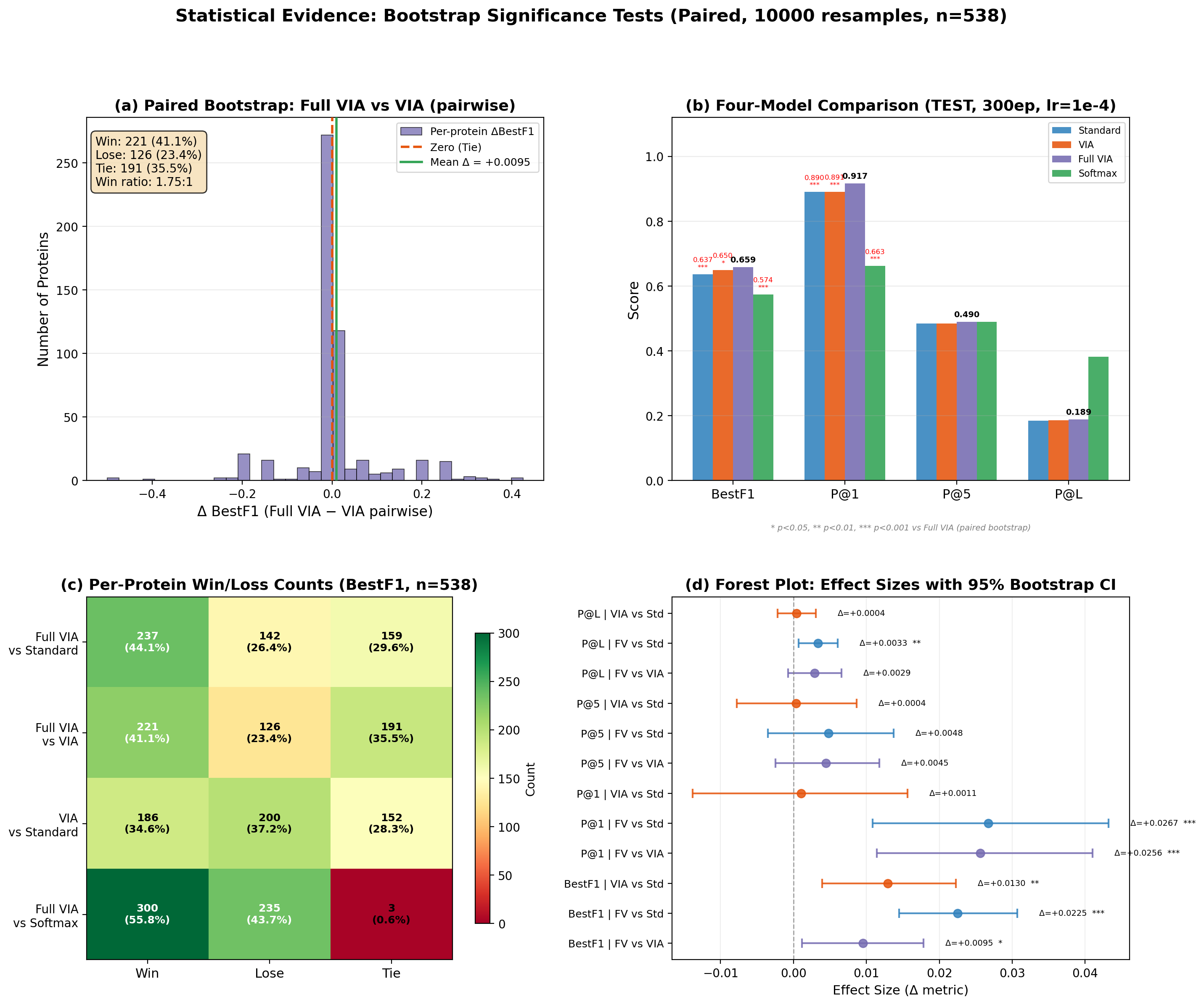}
\caption{Paired bootstrap evidence (300ep, $n=538$). (a) Distribution of $\Delta$BestF1 between FV and VIA. (b) Four-model bar chart with significance markers. (c) Per-protein win/loss/tie heatmap. (d) Forest plot of effect sizes with 95\% bootstrap CI.}
\label{fig:bootstrap-evidence}
\end{figure*}

FV significantly outperforms VIA in BestF1 ($p=0.012$), confirming the incremental value of hierarchical Boltzmann attention. FV achieves highly significant P@1 gains over all baselines ($p<0.001$). VIA (pairwise) significantly outperforms the multi-label baseline in BestF1 ($p=0.002$). Effect sizes are small (Cohen's $d\approx 0.1$), consistent with VIA as a structural refinement. Figure~\ref{fig:bootstrap-evidence} summarizes these results.

\subsubsection*{A3.6 Inference-Time $K$ Ablation}
\label{app:k-ablation}

Using the same FV checkpoint trained with $K=8$, we evaluate $K\in\{1,3,5,7,8\}$ at inference time.

\begin{table}[htbp]
\centering
\setlength{\tabcolsep}{4pt}
\footnotesize
\caption{Inference-time $K$ ablation (protein, same FV checkpoint trained with $K=8$).}
\label{tab:k-ablation}
\begin{tabular}{ccccc}
\toprule
$K$ & BestF1 & P@1 & P@5 & P@L \\
\midrule
1 & $\mathbf{0.6695}$ & 0.9170 & 0.4897 & 0.1885 \\
3 & 0.6653 & 0.9170 & 0.4897 & 0.1885 \\
5 & 0.6638 & 0.9170 & 0.4897 & 0.1885 \\
7 & 0.6599 & 0.9170 & 0.4897 & 0.1885 \\
8 & 0.6591 & 0.9170 & 0.4897 & 0.1885 \\
Standard & 0.6370 & 0.8900 & 0.4850 & 0.1850 \\
\bottomrule
\end{tabular}
\end{table}

\begin{figure*}[t]
\centering
\includegraphics[width=0.7\linewidth]{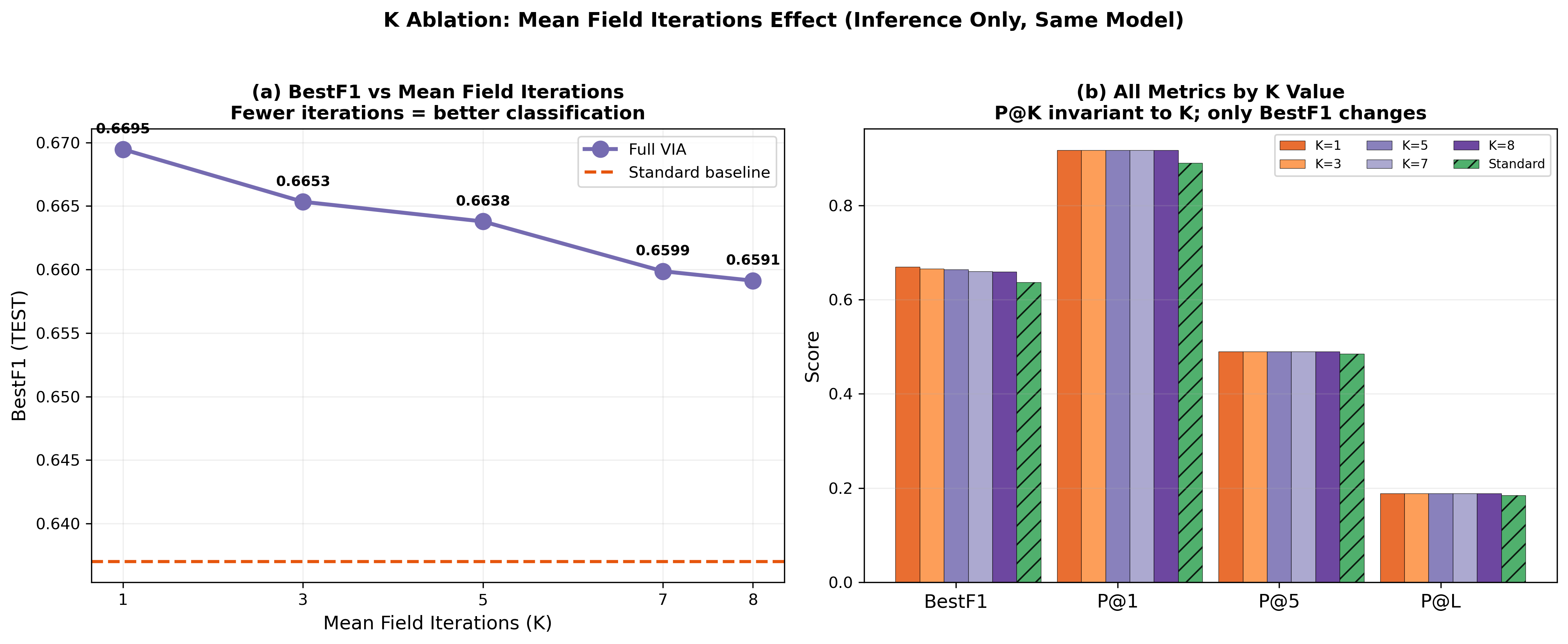}
\caption{Inference-time $K$ ablation. Ranking metrics are invariant to $K$; BestF1 degrades monotonically with more iterations.}
\label{fig:k-ablation}
\end{figure*}

Ranking metrics are \emph{exactly} invariant to $K$: each mean-field update applies a monotone softmax over the same coupled field, rescaling probabilities without changing their order. BestF1 decreases monotonically with $K$, as accumulated coupling pushes probabilities toward extremes and shifts the optimal threshold. Figure~\ref{fig:k-ablation} visualizes this effect.

\subsubsection*{A3.7 J Matrix Structural Analysis}
\label{app:jmatrix}

\paragraph*{Geometric Frustration}

\begin{table}[htbp]
\centering
\setlength{\tabcolsep}{3pt}
\footnotesize
\caption{Triadic sign-pattern counts in the learned $\mathbf{J}$ (protein).}
\label{tab:frustration-full}
\begin{tabular}{cccccc}
\toprule
Pattern & $J_{ij}$ & $J_{jk}$ & $J_{ik}$ & Product & Count \\
\midrule
$+++$ & $+$ & $+$ & $+$ & $+$ (satisfied) & 30{,}296 (30.3\%) \\
$+--$ & $+$ & $-$ & $-$ & $+$ (satisfied) & 36{,}656 (36.7\%) \\
$-+-$ & $-$ & $+$ & $-$ & $+$ (satisfied) & 24{,}900 (24.9\%) \\
$--+$ & $-$ & $-$ & $+$ & $+$ (satisfied) & 8{,}148 (8.1\%) \\
$++-$ & $+$ & $+$ & $-$ & $-$ (frustrated) & 0 \\
$---$ & $-$ & $-$ & $-$ & $-$ (frustrated) & 0 \\
$+-+$ & $+$ & $-$ & $+$ & $-$ (frustrated) & 0 \\
$-++$ & $-$ & $+$ & $+$ & $-$ (frustrated) & 0 \\
\bottomrule
\end{tabular}
\end{table}

Of the couplings, $53.5\%$ are positive and $46.5\%$ negative, yet the Ising frustration ratio is exactly $0.00\%$. All four frustrated patterns have count zero while all four non-frustrated patterns are populated. The learned sign structure is fully satisfiable---there exists a global spin configuration that minimizes all local conflicts. This is the statistical-mechanical signature of a funnel-shaped energy landscape.

\paragraph*{Effective Rank}

\begin{table}[htbp]
\centering
\setlength{\tabcolsep}{4pt}
\footnotesize
\caption{Effective rank and spectral structure of the learned $\mathbf{J}$.}
\label{tab:erank}
\begin{tabular}{lc}
\toprule
Property & Value \\
\midrule
Numerical rank & 25 \\
Effective rank & 2.21 \\
Top-10 singular value energy share & 99.9\% \\
Shannon entropy & 0.79 \\
\bottomrule
\end{tabular}
\end{table}

$\mathbf{J}$'s information is concentrated in 2--3 dimensions despite a numerical rank of 25. The zero frustration ratio is a direct consequence: with effective rank 2.21, the coupling matrix embeds residues into a 2--3 dimensional space where sign conflicts cannot arise. This is consistent with the funnel-shaped energy landscape theory of protein folding.

\paragraph*{Long-Range and Secondary-Structure Coupling}

\begin{table}[htbp]
\centering
\setlength{\tabcolsep}{3pt}
\footnotesize
\caption{Full coupling-strength statistics for the protein model.}
\label{tab:coupling-stats}
\begin{tabular}{p{0.42\columnwidth}ccc}
\toprule
Comparison & $\Delta J$ & Cohen's $d$ & $p$-value \\
\midrule
Global: contact vs.\ non-contact & $+0.776$ & $1.596$ & $<0.001$ \\
Long-range ($|i-j|>24$): contact vs.\ non-contact & $+0.798$ & $1.758$ & $<0.001$ \\
Short-range ($|i-j|\leq 24$): contact vs.\ non-contact & $+0.285$ & $0.752$ & --- \\
Same-SSE vs.\ different-SSE & $+0.752$ & $1.386$ & $<0.001$ \\
\bottomrule
\end{tabular}
\end{table}

The long-range effect size ($d=1.758$) is more than double the short-range effect ($d=0.752$): VIA specifically captures long-range cooperative folding signals rather than recapitulating local proximity. Residues within the same secondary structure element show significantly stronger coupling than those across elements ($d=1.386$).

\begin{figure*}[t]
\centering
\includegraphics[width=\linewidth]{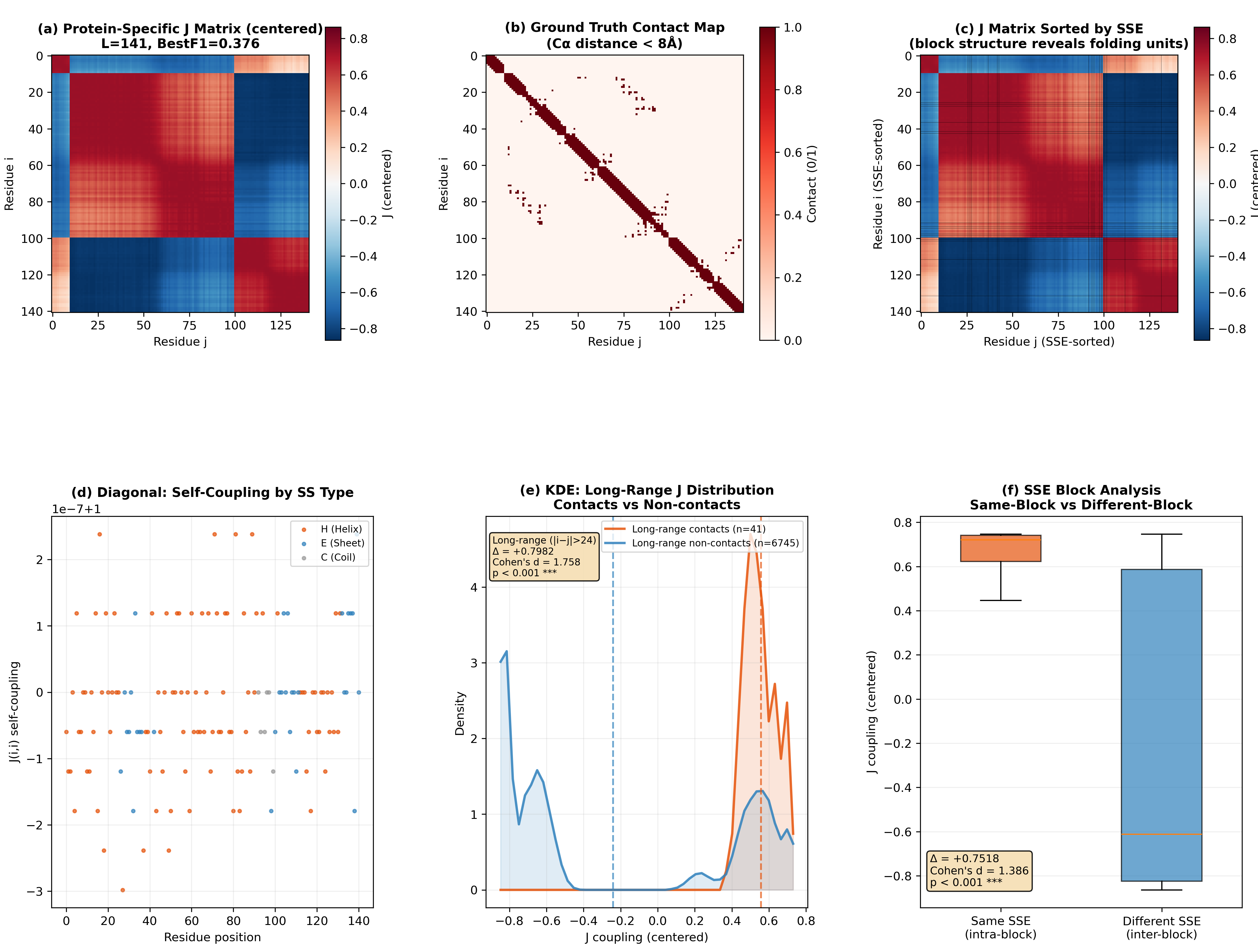}
\caption{Learned coupling matrix $\mathbf{J}$ for a representative protein. (a) Centered $\mathbf{J}$ matrix with block-diagonal structure aligned to secondary structure elements. (b) Ground-truth contact map. (c) $\mathbf{J}$ sorted by SSE. (d) Diagonal self-coupling by SS type. (e) Long-range $J$ distribution: contact vs.\ non-contact. (f) Same-SSE vs.\ different-SSE coupling strengths.}
\label{fig:jmatrix-full}
\end{figure*}

\begin{figure*}[t]
\centering
\includegraphics[width=\linewidth]{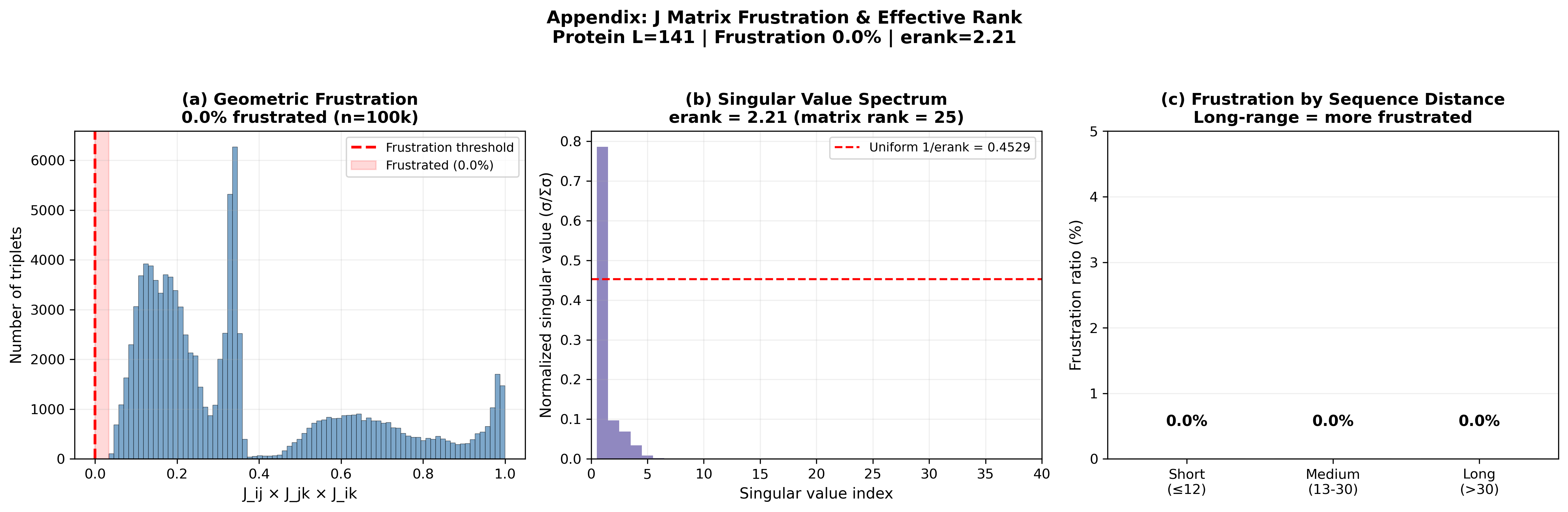}
\caption{J matrix structural analysis. Left: effective rank distribution across variants; right: triadic sign-pattern counts, with all four frustrated patterns at zero.}
\label{fig:frustration-erank}
\end{figure*}

Two figures accompany this analysis. Figure~\ref{fig:jmatrix-full} shows the full coupling matrix for a representative protein (block structure, contact map, SSE-sorted view, diagonal self-coupling, long-range KDE, and SSE block comparison). Figure~\ref{fig:frustration-erank} shows the effective rank distribution and the triadic sign-pattern counts.

\subsubsection*{A3.8 Zero-Feature and Equivariant Controls}

\paragraph{Zero-feature control.} Removing all chemistry features (retaining only masks and slots) collapses every variant to $\leq 17.5\%$ (LowRank: 1.9\%). The slot-indexed coupling stores no standalone position prior; its contribution is multiplicative with chemistry-derived marginals.

\paragraph{Equivariant coupling control.} As a strictly permutation-equivariant parameterization, we use $J_{ij}=\mathrm{MLP}(\mathbf{h}_i\Vert\mathbf{h}_j)$ (no slot-indexed tensors). Its predictions are invariant under atom permutations and canonical reordering, retaining 76.9\% overall and 78.2\% on the shortcut-impossible subset, confirming the mean-field coupling mechanism operates on chemistry features alone.

\subsubsection*{A3.9 Summary of Protein Findings}

Table~\ref{tab:protein-summary} summarizes the main protein-benchmark findings. 
Overall, FV achieves the best performance and consistently outperforms VIA, with statistically significant gains in both BestF1 and P@1, as well as stronger long-range discrimination and a clearer SSE block structure. 
The remaining rows report descriptive diagnostics rather than comparative tests. 
Dashes indicate that no significance test is applicable or reported.

\begin{table}[htbp]
\centering
\setlength{\tabcolsep}{3pt}
\footnotesize
\caption{Summary of protein-benchmark findings.}
\label{tab:protein-summary}
\begin{tabular}{p{0.28\columnwidth}p{0.38\columnwidth}p{0.24\columnwidth}}
\toprule
Finding & Result & Significance \\
\midrule
FV best overall & BestF1 0.659, P@1 0.917 & --- \\
FV vs.\ VIA (BestF1) & $\Delta=+0.010$ & $p=0.012$ \\
FV vs.\ VIA (P@1) & $\Delta=+0.026$ & $p<0.001$ \\
Per-protein win ratio & $221{:}126$ & $1.75{:}1$ \\
Long-range discrimination & Cohen's $d=1.758$ & $p<0.001$ \\
SSE block structure & $\Delta J=+0.752$ & $p<0.001$ \\
Ising frustration & $0.00\%$ & --- \\
Effective rank & 2.21 & --- \\
Optimal inference $K$ & 1 & --- \\
\bottomrule
\end{tabular}
\end{table}

\subsubsection*{A3.10 Figure Index}

Table~\ref{tab:fig-index} lists all figures referenced in this appendix and their purpose.

\begin{table}[htbp]
\centering
\setlength{\tabcolsep}{2pt}
\footnotesize
\caption{Protein-benchmark figures and their purpose.}
\label{tab:fig-index}
\begin{tabular}{p{0.12\columnwidth}p{0.40\columnwidth}p{0.32\columnwidth}}
\toprule
Figure & File & Purpose \\
\midrule
\ref{fig:fair300} & \path{fair_300_comparison.png} & Four-model fair comparison \\
\ref{fig:p1-crossover} & \path{figure_p1_crossover_trajectory.pdf} & 80ep$\to$300ep P@1 crossover \\
\ref{fig:bootstrap-evidence} & \path{figure4_bootstrap_evidence.png} & Bootstrap statistics (4 panels) \\
\ref{fig:k-ablation} & \path{figure_k_ablation.png} & Inference-time $K$ ablation \\
\ref{fig:jmatrix-full} & \path{figure5_J_matrix_visualization.png} & J matrix structure (6 panels) \\
\ref{fig:frustration-erank} & \path{appendix_frustration_erank.png} & Frustration + effective rank \\
\bottomrule
\end{tabular}
\end{table}

\end{document}